\documentclass[11pt]{article}
\def\contents{0}
\def\anonymous{0}

\RequirePackage{amsmath,amssymb,amsthm}
\RequirePackage{tabularx, booktabs, array}
\RequirePackage{bm}
\RequirePackage{dsfont}
\RequirePackage{xcolor}
\RequirePackage{framed}
\RequirePackage{marginnote}
\RequirePackage[skins,breakable]{tcolorbox}
\RequirePackage{algorithm}
\RequirePackage{algorithmic}

\RequirePackage{caption}
\RequirePackage{etoolbox}
\RequirePackage[margin=1.2in]{geometry}
\RequirePackage[bookmarks=false]{hyperref}
\RequirePackage[numbers,square]{natbib}
\let\cite\citep
\let\Cite\citep
\makeatletter
\def\NAT@separator{\NAT@sep\penalty100\relax}
\def\NAT@citea@mbox{\@citea\NAT@hyper@{\citenumfont{\NAT@num}}}

\patchcmd{\@lbibitem}{\item[\hfil}{\item[}{}{}

\@namedef{inlinebibauthor@BrownGSUZ21}{Brown et al.}
\@namedef{inlinebibauthor@AumullerLNP23}{Aumüller et al.}
\@namedef{inlinebibauthor@DaganJYZZ24}{Dagan et al.}
\@namedef{inlinebibauthor@ChenDMM26}{Chen et al.}
\@namedef{inlinebibauthor@SinghalS21}{Singhal and Steinke}
\@namedef{inlinebibauthor@Tsfadia24}{Tsfadia}
\@namedef{inlinebibauthor@BunUV14}{Bun, Ullman, and Vadhan}
\@namedef{inlinebibauthor@DworkSSUV15}{Dwork et al.}
\@namedef{inlinebibauthor@KamathLSU19}{Kamath et al.}
\@namedef{inlinebibauthor@KamathMS22}{Kamath, Mouzakis, and Singhal}
\@namedef{inlinebibauthor@CaiWZ23}{Cai, Wang, and Zhang}
\@namedef{inlinebibauthor@PeterTU24}{Peter, Tsfadia, and Ullman}
\@namedef{inlinebibauthor@SteinkeU17}{Steinke and Ullman}
\DeclareRobustCommand{\citet}[1]{\csname inlinebibauthor@#1\endcsname\space\citep{#1}}
\renewenvironment{thebibliography}[1]
  {\section*{References}%
   \addcontentsline{toc}{section}{References}%
   \normalfont\normalsize
   \list{\@biblabel{\the\c@enumiv}}{%
     \settowidth{\labelwidth}{\@biblabel{#1}}%
     \setlength{\labelsep}{1em}%
     \setlength{\leftmargin}{\labelwidth}%
     \addtolength{\leftmargin}{\labelsep}%
     \setlength{\itemsep}{\bibsep}%
     \setlength{\parsep}{0pt}%
     \usecounter{enumiv}}%
   \let\p@enumiv\@empty
   \interlinepenalty=5000\relax
   \widowpenalty=10000\relax
   \clubpenalty=10000\relax
   \raggedbottom\frenchspacing}
  {\endlist}
\makeatother
\RequirePackage[tt=false]{libertine}
\RequirePackage[libertine]{newtxmath}
\RequirePackage[T1]{fontenc}
\RequirePackage[utf8]{inputenc}
\RequirePackage{microtype}
\RequirePackage{multirow}
\RequirePackage{enumitem}
\RequirePackage{nicefrac}
\RequirePackage{tikz}
\RequirePackage{url}
\RequirePackage{verbatim}
\RequirePackage{wrapfig}
\RequirePackage{nccmath}
\RequirePackage[capitalize]{cleveref}

\usetikzlibrary{positioning}

\definecolor{DarkGreen}{rgb}{0.2,0.6,0.2}
\definecolor{DarkRed}{rgb}{0.6,0.2,0.2}
\definecolor{DarkBlue}{rgb}{0.2,0.2,0.6}
\definecolor{DarkPurple}{rgb}{0.4,0.2,0.4}

\hypersetup{
    linktocpage=true,
    colorlinks=true,
    linkcolor=DarkBlue,
    citecolor=DarkBlue,
    urlcolor=DarkBlue,
}

\definecolor{ProofGray}{gray}{0.25}
\definecolor{ProofRuleGray}{gray}{0.75}
\makeatletter
\renewenvironment{proof}[1][\proofname]{\par
    
    \pushQED{\qed}\normalfont\color{ProofGray}\MakeFramed{\advance\hsize-\width\FrameRestore}\noindent\textit{#1\@addpunct{.}}\enspace\ignorespaces
}{\popQED\endMakeFramed\@endpefalse
}
\makeatother

\newcommand{\norm}[2]{\left \|{#1}\right \|_{#2}}
\newcommand{\abs}[1]{{\left | {#1} \right|}}
\newcommand{\paren}[1]{{\left( {#1} \right)}}
\newcommand{\bparen}[1]{{\big ( {#1} \big)}}

\newcommand{\brackets}[1]{{\left [ {#1} \right]}}

\newcommand{\set}[1]{{\left \{ {#1} \right\}}}

\newcommand{\floor}[1]{\lfloor #1 \rfloor}

\newcommand{\ip}[2]{\left\langle #1, #2\right\rangle}
\newcommand{\ord}[2][th]{\ensuremath{{#2}^{\mathrm{#1}}}}

\let\originalleft\left
\let\originalright\right
\renewcommand{\left}{\mathopen{}\mathclose\bgroup\originalleft}
\renewcommand{\right}{\aftergroup\egroup\originalright}

\DeclareMathOperator*{\Ex}{\mathds{E}}
\DeclareMathOperator*{\Var}{\mathrm{Var}}
\DeclareMathOperator{\proj}{\mathrm{proj}}

\newcommand{\Span}{\mathrm{Span}}

\newcommand{\eps}{\varepsilon}
\newcommand{\id}{\mathbb{I}}
\newcommand{\cond}[1]{\vert_{#1}}

\newcommand{\nicehalf}{\nicefrac{1}{2}}

\newcommand{\invchi}{\mathsf{inv\chi^2}}

\newcommand{\Ind}{\mathds{1}}
\newcommand{\tr}{\mathrm{tr}}

\newcommand{\risk}{\mathfrak{R}}

\newcommand{\prisk}{\risk_{\eps,\delta}}
\newcommand{\pkrisk}{\risk^{\text{\tiny known}}_{\eps,\delta}}

\newcommand{\N}{\mathbb{N}}
\newcommand{\R}{\mathbb{R}}

\newcommand{\cA}{\mathcal{A}}

\newcommand{\cH}{\mathcal{H}}

\newcommand{\cM}{\mathcal{M}}
\newcommand{\cN}{\mathcal{N}}

\newcommand{\cP}{\mathcal{P}}
\newcommand{\cQ}{\mathcal{Q}}

\newcommand{\cT}{\mathcal{T}}

\newcommand{\cW}{\mathcal{W}}
\newcommand{\cX}{\mathcal{X}}

\newcommand{\rv}[1]{\bm{#1}}

\newcommand{\rvD}{\rv{D}}

\newcommand{\rvJ}{\rv{J}}

\newcommand{\rvM}{\rv{M}}

\newcommand{\rvS}{\rv{S}}
\newcommand{\rvT}{\rv{T}}

\newcommand{\rvX}{\rv{X}}
\newcommand{\rvY}{\rv{Y}}
\newcommand{\rvZ}{\rv{Z}}

\newcommand{\rvg}{\rv{g}}

\newcommand{\rvi}{\rv{i}}

\newcommand{\rvk}{\rv{k}}

\newcommand{\rvq}{\rv{q}}
\newcommand{\rvr}{\rv{r}}
\newcommand{\rvs}{\rv{s}}

\newcommand{\rvu}{\rv{u}}
\newcommand{\rvv}{\rv{v}}
\newcommand{\rvw}{\rv{w}}
\newcommand{\rvx}{\rv{x}}
\newcommand{\rvy}{\rv{y}}
\newcommand{\rvz}{\rv{z}}

\newcommand{\rvSigma}{\rv{\Sigma}}
\newcommand{\rvsigma}{\rv{\sigma}}
\newcommand{\rvmu}{\rv{\mu}}
\newcommand{\rvpi}{\rv{\pi}}
\newcommand{\rvtheta}{\rv{\theta}}

\newtheorem{thm}{Theorem}[section]
\newtheorem*{thm*}{Theorem}

\newtheorem{claim}[thm]{Claim}

\newtheorem{theorem}[thm]{Theorem}
\newtheorem{lemma}[thm]{Lemma}

\newtheorem{fact}[thm]{Fact}
\newtheorem{proposition}[thm]{Proposition}

\theoremstyle{definition}

\newtheorem{definition}[thm]{Definition}

\crefname{claim}{claim}{claims}
\Crefname{claim}{Claim}{Claims}
\crefname{thm}{theorem}{theorems}
\Crefname{thm}{Theorem}{Theorems}
\crefname{fact}{fact}{facts}
\Crefname{fact}{Fact}{Facts}

\newtheoremstyle{boldremark}
  {.5\topsep}{.5\topsep}
  {\normalfont}{}
  {\em\bfseries}{.}{5pt plus 1pt minus 1pt}{}
\theoremstyle{boldremark}
\newtheorem{remark}{Remark}

\iftrue

\else

\fi
\title{Dimension-Free Private Mean Estimation is Impossible Without the Covariance}

\ifnum\anonymous=1
    \author{Anonymous Author(s)}
\else
    \author{
        Renaud Gaucher\thanks{Center of Applied Mathematics, École polytechnique. This work was partially conducted while the author was at UC Berkeley. Supported by French State managed by the ANR under France 2030 program with the reference ANR-23-PEIA-005 (REDEEM project) and by a public grant from the Fondation Mathématique Jacques Hadamard.}~~~~~~~~ 
        \and Ephraim Linder\thanks{Department of Computer Science, Boston University.}~~~~~~~~
        \and Adam Smith\thanks{Department of Computer Science, Boston University. This work was partially supported by NSF award CNS-2232694 and an Apple faculty research award.}
        \and Jonathan Ullman\thanks{Khoury College of Computer Sciences, Northeastern University. Supported by NSF CNS-2232692 and CNS-2247484.}~~~~~~~~
        \and Lydia Zakynthinou\thanks{Department of Computer Science, Johns Hopkins University. This work was partially conducted while the author was at The Simons Institute for the Theory of Computing at UC Berkeley.}
    }
    \date{\today}
\fi

\begin{document}
\pagenumbering{roman}
\maketitle

\begin{abstract}

We study the limits of differentially private Gaussian mean estimation when the covariance matrix is unknown. Prior work shows that when the covariance $\Sigma$ is known, the optimal sample complexity can depend only on its spectrum, rather than on the ambient dimension $d$. We establish new lower bounds for Gaussian mean estimation with Euclidean error which show that private estimators cannot achieve similar guarantees when the covariance is unknown. First, when the sample size $n$ is smaller than the effective dimension $\mathrm{trace}(\Sigma)/\|\Sigma\|_{\text{op}}$, the optimal error when the covariance is unknown can be larger than the optimal error when it is known by a factor of $\sqrt{d/n}$. Second, when the sample size is larger than the effective dimension, even for diagonal covariance matrices with effective dimension bounded by $\log d$, the optimal error when the covariance is unknown can be larger than the optimal error when it is known by a factor of $\sqrt{d}/n^{3/2}$. Together, these results show that in the unknown covariance setting, exploiting anisotropic structure under differential privacy can require a dimension-dependent error or sample complexity.
\end{abstract}

\ifnum\contents=1
\newpage
\tableofcontents
\fi

\newpage \pagenumbering{arabic}
\section{Introduction}

Statistical inference and machine learning algorithms are designed to extract population-level insights from data, but in doing so, they routinely reveal information about the individuals whose data they process. The risks of privacy violations are well documented, ranging from reconstruction~\citep{DinurN03, DworkMT07, DworkY08, KasiviswanathanRSU10}, membership-inference~\citep{Homer+08, SankararamanOJH09, BunUV14, DworkSSUV15, ShokriSSS17, YeomGFJ18} and the unintended memorization of training data~\citep{CarliniLEKS19, Carlini+21, Feldman20, BrownBFST21}. 
Differential privacy (DP)~\citep{DworkMNS06} provides a rigorous framework for mitigating these risks by limiting the effect of any individual’s data on the distribution of an algorithm’s output, and has become a standard approach to privacy protection in statistics and machine learning. 

At the same time, ensuring privacy can come at a substantial cost in statistical accuracy, 
especially in high-dimensional settings. 
This has motivated an extensive literature characterizing the tradeoffs between privacy, sample size, and accuracy, 
across a range of high-dimensional statistical and machine learning tasks. 
A central question is whether private algorithms can exploit structure in the underlying distribution to reduce this cost. 

\paragraph{Gaussian mean estimation.}
A natural setting in which to study these tradeoffs is Gaussian mean estimation. 
Given independent samples from a Gaussian distribution with unknown mean $\mu\in\R^d$ and covariance matrix $\Sigma$, the goal is to estimate $\mu$ accurately in Euclidean distance. Mean estimation is a fundamental statistical task in its own right and serves as a building block for more complex procedures, including regression and optimization. Its simplicity also makes it an ideal setting in which we can isolate the interplay between privacy, dimensionality, and covariance structure.

\paragraph{Summary of our contributions.} In this work, we compare the optimal error achievable for private Gaussian mean estimation when the covariance is unknown to the optimal error when the covariance is known (i.e., given as input to the algorithm). In the known covariance setting, recent work \cite{DaganJYZZ24} shows that the optimal Euclidean error scales as 
\begin{align}
    \label{eq:error-known}
    \tilde\Theta\left(\sqrt{\frac{\tr(\Sigma)}{n}} + \frac{\tr\left(\Sigma^{1/2}\right)\sqrt{\log(1/\delta)}}{\eps n}\right),
\end{align}
where the first term is the statistical rate of Gaussian mean estimation, and the second term is the additional error due to privacy. Remarkably, the bound achieved by \cite{DaganJYZZ24} is ``dimension-free''---the error depends only on the geometry of the data, and is independent of the ambient dimension $d$. Our main contributions show that this dimension-free behavior cannot be recovered in the unknown covariance setting. Our first contribution shows that for a sample of size $n$, the optimal error in the unknown covariance setting can be larger by a factor of $\sqrt{d/n}$ than the optimal error in the known covariance setting. Our second contribution considers the setting when the covariance is promised to be diagonal: we show that even under this additional promise, the optimal error in the unknown covariance setting can be larger by a factor of $\sqrt{d}/n^{1.5}$ than the optimal error in the known covariance setting. \Cref{tab:results-informal} summarizes our main contributions. 

\begin{table}[H]
\centering
\footnotesize
\renewcommand{\arraystretch}{2.5}
\begin{tabular}{c|c|c}
\hline
Setting & Error blowup & Obstruction \\
\hline
General covariance 
& 
${\displaystyle
\mathrm{error}_{\mathrm{unknown}}=\tilde{\Theta}_{\eps,\delta}\left(1+\sqrt{\frac{d}{n}}\right)\cdot\mathrm{error}_{\mathrm{known}}
}$ 
& 
Subspace estimation\\
\hline 
Diagonal covariance
& 
${\displaystyle
\mathrm{error}_{\mathrm{unknown}}=\tilde{\Theta}_{\eps,\delta}\left(1+\frac{\sqrt{d}}{n^{1.5}}\right)\cdot\mathrm{error}_{\mathrm{known}}
}$
& 
Identifying large variance coordinates\\
\end{tabular}
\begin{minipage}{\textwidth}
\caption{$\mathrm{error}_{\mathrm{unknown}}$ and $\mathrm{error}_{\mathrm{known}}$, respectively, denote the Euclidean error in the unknown and known covariance settings. The known covariance bound, and the upper bound for unknown diagonal covariance are due to \cite{DaganJYZZ24}. The upper bound for unknown general covariance is a folklore algorithm based on the framework from \cite{TsfadiaCKMS22} and presented in \Cref{sec:folklore-estimator}. The lower bounds for $\mathrm{error}_{\mathrm{unknown}}$ in both settings are \Cref{thm:diag:estimation-lb,thm:nd:estimation-lower-bound} of this work.}
\label{tab:results-informal}
\end{minipage}
\end{table}

Our bounds highlight two fundamental challenges for mean estimation with unknown covariance. In the promised diagonal setting, we show that every private mean estimation algorithm must, in some sense, ``identify'' many of the coordinates with the largest variances. And, in the general covariance setting, we show that every private mean estimation algorithm must learn the subspace spanned by the covariance. 

In addition to our lower bounds for mean estimation in Euclidean distance, we show that in the more challenging setting of estimation in Mahalanobis error---where the error of the estimator in each direction is required to be proportional to the variance in that direction---private mean estimation with unknown covariance is impossible whenever $n\leq d-2$. This is in stark contrast to the known covariance setting where the optimal Mahalanobis error is finite at essentially all sample sizes.    

\subsection{Our results}

All of our lower bounds apply to the approximate DP setting defined below. 

\begin{definition}[Differential Privacy \citep{DworkMNS06}]
    A randomized algorithm $\cA:\cX^* \rightarrow \cW$ is $(\eps,\delta)$-differentially private, if, for all neighboring datasets $X, X'$,i.e. datasets that differ on a single entry, and all measurable subsets $W\subset \cW$ we have 
    $$
    \Pr[\cA(X)\in W] \leq e^{\eps}\Pr[\cA(X')\in W] + \delta.
    $$
\end{definition}

In order to facilitate comparison between the error in the known and unknown covariance settings, and to present a scale-free ``cost'' of not knowing the covariance, we define the known and unknown covariance Bayes risk. 

\begin{definition}[Bayes risk]
\label{def:bayes_risk}
    Let $A_{\eps,\delta}$ denote the set of $(\eps,\delta)$-DP algorithms, and let $\cP$ be a prior distribution over $\rvmu \in \R^d$ and positive semi-definite $\rvSigma \in \R^{d\times d}$. The \emph{unknown-covariance Bayes risk} with respect to the prior $\cP$ is defined as:
    \[
    \prisk\paren{\cP,n} = \inf\left\{ \alpha \ge 0: \exists\, \cA\in A_{\eps,\delta} ~~\text{s.t.}~ \Pr_{\substack{(\rvmu,\rvSigma)\sim\cP \\ \rvX\sim \cN(\rvmu,\rvSigma)^{\otimes n}}}\brackets{\norm{\cA(\rvX) - \rvmu}{2}\leq\alpha}\geq \frac 34 \right\}.
    \]
We also define the \emph{known-covariance Bayes risk} (i.e., the risk when the algorithm is given the covariance as input) with respect to the prior $\cP$ by:
    \[
    \pkrisk\paren{\cP,n} = \inf\left\{\alpha \ge 0: \exists\, \cA\in A_{\eps,\delta} ~~\text{s.t.}~\Pr_{\substack{(\rvmu,\rvSigma)\sim\cP \\ \rvX\sim \cN(\rvmu,\rvSigma)^{\otimes n}}}\brackets{\norm{\cA(\rvX; \rvSigma) - \rvmu}{2}\leq\alpha}\geq \frac 34\right\}.
    \]
\end{definition}

Our results demonstrate the impossibility of private mean estimation with both dimension-free error and dimension-free sample complexity in two settings. Each setting highlights a distinct aspect of private mean estimation that requires many samples: In the first setting, the variance in each basis direction is fixed, and the difficulty stems from the fact that the basis directions are chosen uniformly at random, while in the second setting, the basis directions are fixed, and the difficulty stems from the fact that the variance in each direction is chosen from a suitably pathological distribution. 

\paragraph{General covariance.} In this  setting, we consider mean estimation when the mean $\rvmu$ and the covariance $\rvSigma$ are sampled from the prior $\cP_{\text{sub}}$ defined by 
\begin{align*}
    \paren{\rvmu,\rvSigma}\sim\cP_{\text{sub}}(d,k) \iff 
    \begin{cases}
        \rvmu\sim \cN(0,d^2 \id_d) \\\rvSigma = \rvD\rvD^\top \ \text{where}\ 
        \rvD\sim \cH_{d,k}.
    \end{cases}
\end{align*}
where $\cH_{d,k}$ is the uniform distribution 
over $d\times k$ matrices with orthonormal columns. Note that the effective dimension of every matrix in the support of the prior is $k$, and that the resulting distributions $\cN(\rvmu,\rvSigma)$ are highly structured---the data are isotropic in a $k$-dimensional subspace. Indeed, if $\rvSigma$ is known, then the problem reduces to Gaussian mean estimation with covariance $\id_k$. Using the bound in \eqref{eq:error-known} for the error in the known covariance setting yields\begin{align}
\label{eq:intro:nd:risk-known}
\pkrisk(\cP_{\text{sub}},n)=\tilde O_{\eps,\delta}\paren{\sqrt{\frac{k}{n}} + \frac{k}{n}}
\end{align}
On the other hand, when $\rvSigma$ is unknown, a folklore algorithm based on the FriendlyCore framework of \Cite{TsfadiaCKMS22} (presented in \Cref{sec:folklore-estimator} for completeness) bounds the unknown covariance risk by $\tilde{O}_{\eps,\delta}\left(\sqrt{\frac{\tr(\Sigma)}{n}} + \frac{\sqrt{d\tr(\Sigma)}}{n}\right)$, which yields
\begin{align}
\label{eq:intro:nd:risk-unknown}
\prisk(\cP_{\text{sub}},n)=\tilde O_{\eps,\delta}\paren{\sqrt{\frac{k}{n}} + \frac{\sqrt{dk}}{n}}.
\end{align}

Our main result shows that when $n\ll k$ the folklore estimator is optimal---in particular, a private algorithm cannot take advantage of the low dimensional structure until it has seen almost every direction in the span of $\rvSigma$. 

\medskip
\noindent{\parbox{\textwidth}{
\textbf{\Cref{thm:nd:estimation-lower-bound}.}~ \itshape{For all $n,k,d\in\N$ with $k\lesssim d$ and $n\log n\lesssim k$, if $\eps\leq 1$ and $\delta\lesssim 1/n$ then every $(\eps,\delta)$-DP algorithm $\cM$ satisfies
$$
\Pr_{\substack{(\rvmu,\rvSigma)\sim\cP_{\text{sub}} \\ \rvX\sim\cN(\rvmu,\rvSigma)^{\otimes n}}}\brackets{\|\cM(\rvX)-\rvmu\|_2\geq \frac{\sqrt{dk}}{n}}\geq \frac{1}{4}.
$$
In particular, $\prisk(\cP_{\text{sub}}, n)\geq \frac{\sqrt{dk}}{n}$. 
}}}
\par\medskip
\Cref{thm:nd:estimation-lower-bound} is tight in two respects: First, it matches the upper bound in \eqref{eq:intro:nd:risk-unknown} for the setting where $n\log n < k$, and second, once $n\gtrsim k+\eps^{-1}$, one can use the results of \cite{SinghalS21} to learn $\rvSigma$ exactly and recover the bound in \eqref{eq:intro:nd:risk-known}. Additionally, for the setting of $n\log n < k$, comparing the bound in \Cref{thm:nd:estimation-lower-bound} and the bound in \eqref{eq:intro:nd:risk-known} yields the following scale-free bound on the cost of unknown covariance in private mean estimation,
$$
\prisk\paren{\cP_{\text{sub}},n} = \tilde\Omega_{\eps,\delta}\paren{1+\sqrt{\frac{d}{k}}~}\cdot \pkrisk\paren{\cP_{\text{sub}},n}.
$$
Setting $k\approx n\log n$ recovers the blowup for general covariance in \Cref{tab:results-informal}.

\paragraph{Diagonal covariance.} While the previous setting leverages the randomness in the basis vectors, in the second setting, we instead fix the basis vectors (i.e., we consider diagonal covariance), and leverage the randomness in the variances themselves. To do so, we consider the prior $\cP_{\text{diag}}$ defined by
\begin{align*}
    (\rvmu,\rvSigma) \sim \cP_{\text{diag}}(d) \iff 
    \begin{cases}
        \rvmu \sim \cN(0, d^2\id_d)\\
        \rvSigma = \mathrm{Diag}((\rvsigma^2_i)_{i\in[d]}), ~ \text{where} ~   (\rvsigma^2_i)_{i\in[d]} \sim \invchi(1,1)^{\otimes d}
    \end{cases}
\end{align*}
where $\invchi(\nu,\tau^2)$ denotes the scaled-inverse chi-squared distribution with $\nu$ degrees of freedom and scale $\tau$ (see \Cref{def:inv-chi}). One can show that under the prior $\cP_{\text{diag}}$ the covariance $\rvSigma$ satisfies $\tr(\rvSigma)\approx d^2$ and $\tr(\rvSigma^{1/2})\approx d\log d$ with high constant probability (this can be formalized using standard bounds for sums of heavy tailed random variables, see \Cref{lem:heavy-tailed-sum}). 

For the unknown diagonal covariance setting, an algorithm from \cite{DaganJYZZ24} gives the general bound $\tilde{O}_{\eps,\delta}\left(\sqrt{\frac{\tr(\Sigma)}{n}}
+
\left(1+\frac{\sqrt{d}}{n}\right) \frac{\tr(\Sigma^{1/2})}{n}\right)$, which combined with the bound \eqref{eq:error-known} for private estimation with known covariance, yields the known and unknown covariance risk bounds for $\cP_{\text{diag}}$:
\begin{align}
    \label{eq:intro:diag:risk}
    \pkrisk\paren{\cP_{\text{diag}},n} = \tilde O_{\eps,\delta}\paren{\frac{d}{\sqrt{n}}} \qquad \text{and} \qquad  \prisk\paren{\cP_{\text{diag}},n} = \tilde O_{\eps,\delta}\paren{\frac{d}{\sqrt{n}} + \frac{d^{3/2}}{n^2}}.
\end{align}

Our main result shows a tight lower bound on the unknown-covariance risk for the prior $\cP_{\text{diag}}$.    

\medskip
\noindent{\parbox{\textwidth}{
\textbf{\Cref{thm:diag:estimation-lb}.}~ \itshape{For all $n,d\in\N$ with $n\log n\lesssim\sqrt{d}$, if $\eps\leq 1$ and $\delta\lesssim 1/n$ then every $(\eps,\delta)$-DP algorithm $\cM$ satisfies
$$
\Pr_{\substack{(\rvmu,\rvSigma)\sim\cP_{\text{diag}} \\ \rvX\sim\cN(\rvmu,\rvSigma)^{\otimes n}}}\brackets{\|\cM(\rvX)-\rvmu\|_2\geq \frac{d^{3/2}}{n^2\log d}}\geq \frac{1}{4}.
$$
In particular, $\prisk\paren{\cP_{\text{diag}},n}\geq \frac{d^{3/2}}{n^2}$. 
}}}
\par\medskip
As a corollary of \Cref{thm:diag:estimation-lb} and \eqref{eq:intro:diag:risk}, we obtain the scale-free lower bound on the cost of unknown diagonal covariance in private mean estimation
$$
\prisk\paren{\cP_{\text{diag}},n} =  \tilde \Omega_{\eps,\delta}\paren{1+\frac{\sqrt{d}}{n^{3/2}}}\cdot \pkrisk\paren{\cP_{\text{diag}},n}.
$$
While the blowup is $\Theta(1)$ once $n\geq d^{1/3}$, we note that the bound in \Cref{thm:diag:estimation-lb} nevertheless establishes the necessity of the additive $d^{3/2}/n^2$ term for the full regime where $n\log n\lesssim \sqrt{d}$. 

\Cref{tab:lower-bounds} summarizes our lower bounds for estimation with Euclidean error and includes the matching upper bounds from prior work.

\begin{table}[H]
\centering
\footnotesize
\renewcommand{\arraystretch}{3}
\begin{tabular}{c|c|c}
\hline
Covariance family & Upper bound (prior work) & Lower bound (this work) \\
\hline
Random rank-$k$ orthogonal projector
& 
$\displaystyle
  \tilde O_{\eps,\delta}\left(
    \sqrt{\frac{k}{n}}
    +\frac{\sqrt{dk}}{n}
  \right)$ 
\citep{TsfadiaCKMS22}
& 
$\displaystyle
  \Omega\left(\frac{\sqrt{dk}}{n}\right)$ ~for
  $n\log n \lesssim k$ ~(\Cref{thm:nd:estimation-lower-bound})\\
\hline
Inverse chi-squared diagonal prior
& 
$\displaystyle
  \tilde{O}_{\eps,\delta}\Bigg(\frac{d}{\sqrt n}
    +\frac{d^{3/2}}{n^2}
  \Bigg)$
  \cite{DaganJYZZ24}
& 
$\displaystyle
  \Omega\left(
    \frac{d^{3/2}}{n^2}
  \right)$ ~for
  $n\log n\lesssim \sqrt{d}$
    ~(\Cref{thm:diag:estimation-lb})\\
\end{tabular}
\begin{minipage}{\textwidth}
\caption{Euclidean upper and lower bounds for the two hard covariance priors.
We assume $\eps\leq 1$ and $\delta\lesssim 1/n$. All bounds hold with constant probability for the corresponding covariance priors.}
\label{tab:lower-bounds}
\end{minipage}
\end{table}

\paragraph{Estimation in Mahalanobis norm.} In addition to studying private mean estimation with error in $\ell_2$ norm, we show strong impossibility results for private mean estimation with error measured in Mahalanobis norm using $n\ll d$ samples. For a positive semi-definite matrix $\Sigma$ and a vector $y$, we let $d_{\Sigma}(y)$ denote the Mahalanobis norm of $y$ (defined in \Cref{sec:rel}). Our main result says that for $n\leq d-2$, every $(\eps,\delta)$-DP algorithm achieves finite Mahalanobis error with probability at most $n\delta$.

\medskip
\noindent{\parbox{\textwidth}{
\textbf{\Cref{thm:rel:estimation-lb}.}~ \itshape{For all $d\in\N$ there exists a prior $\cP_{\text{rel}}$ over instances $(\rvmu,\rvSigma)$ such that if $n\leq d-2$, then for all $\eps,\delta>0$ every $(\eps,\delta)$-DP algorithm $\cM$ satisfies
$$
\Pr_{\substack{(\rvmu,\rvSigma)\sim\cP_{\text{rel}} \\ \rvX\sim\cN(\rvmu,\rvSigma)^{\otimes n}}}\brackets{d_{\rvSigma}\paren{\cM(\rvX)-\rvmu} < \infty}\leq n\delta.
$$
}}}
\par\medskip
For $\delta<\frac{1}{n}$, \Cref{thm:rel:estimation-lb} shows that private mean estimation with finite Mahalanobis error is essentially impossible with fewer than $d$ samples. This is in stark contrast to the known covariance setting where finite Mahalanobis error can be achieved for essentially any $n$. For $\delta \geq 1/n$ a ``name-and-shame'' algorithm that publishes $\floor{\delta n}$ records selected randomly from the dataset achieves finite Mahalanobis error, but fails most intuitive concepts of privacy.

\subsection{Our techniques}

In contrast to existing lower bounds for private Gaussian mean estimation, which proceed by sampling $\rvmu$ from a suitable prior and sampling the data from $\cN(\rvmu,\id_d)$, to prove lower bounds in our unknown covariance setting we must design a prior over both $\rvmu$ and $\rvSigma$. Intuitively, in order to show a lower bound that exceeds the lower bound in the known covariance setting, it suffices to show that an estimate of $\rvmu$ in the unknown covariance setting also contains some non-trivial information about $\rvSigma$, and that learning this information privately requires many samples. Indeed, a key step in our proofs is a reduction from a task $T(\rvSigma)$ that only involves $\rvSigma$---that is, we leverage the randomness in $\rvmu$ to show that a mechanism which accurately estimates $\rvmu$ can be used to construct an algorithm which solves the task $T(\rvSigma)$. We then prove a lower bound for privately solving the task $T(\rvSigma)$, which by the reduction, implies a lower bound for privately estimating $\rvmu$. 

\paragraph{General covariance.} To prove our lower bound for general covariance, we define the prior over $\rvSigma$ as follows: sample a uniformly random orthonormal $d\times k$ matrix $\rvD$ and let $\rvSigma=\rvD\rvD^\top$---that is, $\rvSigma$ is a projection matrix onto a uniformly random $k$-dimensional subspace. The key idea used to prove the bound is that for $n\leq k/2$, conditioned on $\rvmu$ and the sample $\rvX=\rvx_1,\dots,\rvx_n\sim\cN(\rvmu,\rvSigma)$, the remaining $k-n$ unseen directions in $\Span(\rvSigma)$ are uniformly random in the remaining $d-n$ dimensional space. Intuitively, a private mechanism must add enough noise to hide the contribution of a single sample, and thus it must add noise in every direction in $\Span(\rvSigma)$; however, since conditioned on the sample, the remaining directions in $\Span(\rvSigma)$ are uniformly random in the ambient space, a private algorithm must add noise in every direction, which results in dimension-dependent error.

We formalize this intuitive argument by showing that every private and accurate mean estimation algorithm can be used to privately generate a sample that simultaneously ``correlates'' with the covariance $\rvSigma$ and has small $\ell_2$ norm. We call this sampling problem $(\lambda,\alpha)$-sampling, where $\lambda$ controls the ``correlation'' between $\rvy$ and $\rvSigma$, namely $\Ex[\rvy^\top\rvSigma\rvy]\geq \lambda^2$, and $\alpha$ controls the norm, that is $\norm{\rvy}{2}\leq \alpha$. Essentially, our reduction shows that every private mean estimation algorithm with error $\alpha$ and sample complexity $n$ can be used to construct a private $(k/n,\alpha)$-sampler with sample complexity $n$. To complete the proof, we show that every private $(\lambda,\alpha)$-sampler with sample complexity $n\leq k/2$ requires $\alpha\geq \lambda\sqrt{d/k}$. Plugging in the parameters from the reduction yields that every private mean estimation algorithm with sample complexity $n\leq k/2$ has error
$$
\alpha\geq \frac{\sqrt{dk}}{n}.
$$

\subparagraph{Sampler lower bound.} To prove a lower bound for private sampling, we design a membership inference attack which uses a test that directly measures the correlation between the target point and the sample output by the mechanism---that is, for test point $z\in\R^d$, sample $y\in\R^d$, and covariance $\Sigma$, we define the test
\begin{align*}
    \cT_{y,\Sigma}(z) = \frac{\langle z, y\rangle^2}{y^\top\Sigma y}.
\end{align*}
Intuitively, if a sampler has seen less than $k/2$ samples, then in order to make $\rvy^\top\rvSigma\rvy$ large, it must either sample from a distribution with large variance in all directions, or it must make $\rvy$ large in the directions it has seen, which are highly dependent on the sample $\rvX$. Since the output of a sampler must satisfy $\norm{\rvy}{2}\leq \alpha$, the former option is impossible, and thus, we expect the output $\rvy$ to be highly dependent on the sample $\rvX$.  
Somewhat more formally, suppose that we sample $\rvSigma$ as follows: first sample the data $\rvx_1,\ldots,\rvx_n$ and let $P_{\rvX}$ denote the projector on $\Span(\rvX)$. Second, since $P_{\rvX}$ defines an $n$ dimensional subspace, we can sample $\rvSigma$ by sampling the remaining $k-n$ directions uniformly at random from its orthogonal complement. We can then decompose the correlation between $\rvy$ and $\rvSigma$ into the correlation between $\rvy$ and $P_{\rvX}$, and the correlation between $\rvy$ and a uniformly random $k-n$ dimensional projection matrix: 
$$
\rvy^\top\rvSigma\rvy = \rvy^\top P_{\rvX}\rvy + \rvy^\top\paren{\rvSigma - P_{\rvX}}\rvy,
$$
where $\paren{\rvSigma - P_{\rvX}}$ is a uniformly random projection from a $d-n$ dimensional space to a $k-n$ dimensional space. Thus using the norm bound on $\rvy$, we have that $\rvy^\top\paren{\rvSigma - P_{\rvX}}\rvy$ is typically at most $\alpha^2 k/d$. Since the left hand side ($\rvy^\top\rvSigma\rvy$) is in expectation at least $\lambda^2$, we see that $\rvy^\top P_{\rvX}\rvy\geq \lambda^2/2$ whenever $\alpha\leq \lambda\sqrt{d/2k}$. Thus, any $(\lambda,\alpha)$-sampler with $\alpha\leq \lambda\sqrt{d/2k}$ must exhibit significant correlation with the data $\rvX$. Leveraging this correlation allows us to show that the attack $\cT$ can distinguish a member in $\rvX$ from a fresh sample from the population, and thus, such a $(\lambda,\alpha)$-sampler cannot be private. 

\paragraph{Diagonal covariance.} A first step in proving the lower bound in the diagonal covariance setting is constructing a prior over $\rvSigma$ (or family of covariance matrices) for which we expect private mean estimation with unknown covariance to be much harder than private estimation with known covariance. Our starting point is the algorithm of \cite{DaganJYZZ24} which proceeds as follows: First learn the coordinates with the $k\approx n^2$ largest variances, and estimate the mean in those coordinates using the known covariance algorithm; and second, estimate the mean in the remaining $d-k$ coordinates by adding spherical noise proportional to the trace. At a high level, one can show that the relative error of this algorithm compared to the optimal known covariance algorithm is largest when the diagonal entries $\rvsigma^2_i$ decay proportional to $i^{-2}$. This leads us to sample each $\rvsigma^2_i$ independently from the scaled-inverse-chi-squared distribution $\invchi(1,1)$. This distribution has a tail of order $t^{-1/2}$, so its $i$-th largest draw is typically on the scale $d^2i^{-2}$.

Next, to prove our lower bound for diagonal covariance, we show that every private and accurate mean estimation algorithm can be used to privately select the coordinates with the $k$ largest variances---that is, we reduce from a ``top-$k$'' variance selection problem that we call $(k,\ell)$-selection. In $(k,\ell)$-selection we are given samples $\rvX=\rvx_1,\dots,\rvx_n\sim\cN(0,\rvSigma)^{\otimes n}$ where $\rvSigma=\mathrm{Diag}(\rvsigma^2_1,\dots,\rvsigma^2_d)\sim\invchi(1,1)^{\otimes d}$, and the goal is to output a list of at most $\ell$ coordinates that contains at least $1/4$ of the coordinates with the $k$-largest variances. Formally, our reduction shows that every private mean estimation algorithm that achieves error $\alpha$ can be used to privately select the coordinates with the $k$ largest variances, while selecting at most $\ell\approx \frac{\alpha^2 k n^2}{d^2}$ coordinates overall. We complete the proof by showing a lower bound for private $(k,\ell)$ selection which states that any private $(k,\ell)$-selection algorithm with $n\leq \sqrt{k}$ samples requires $\ell\geq d/\log d$. Setting $k\approx n^2$ and combining the selection lower bound with the setting of $\ell$ in the reduction, we obtain that every private mean estimation algorithm has error 
$$
\alpha\geq \frac{d^{3/2}}{n\sqrt{k}}\approx \frac{d^{3/2}}{n^2}.
$$

\subparagraph{Selection lower bound.} To show a lower bound for private selection we construct a membership inference attack against any $(k,\ell)$-selection algorithm that uses $n\leq \sqrt{k}$ samples and at the same time selects at most $\ell\leq d/\log d$ coordinates. One natural candidate for such an attack, inspired by \cite{SteinkeU17} is the following test: For test point $z\in\R^d$, covariance $\Sigma$, and selection vector $s\in\{0,1\}^d$, define 
\begin{align*}
    \cT_{\text{naive}}(z,s,\Sigma) := \sum_{j \in [d]}\paren{\frac{z_j^2}{\sigma_j^2}-1} s_j.
\end{align*}
One might hope that when the test point $\rvx_{\rvi}$ is in the dataset the expectation $\Ex[\cT_{\text{naive}}(\rvx_{\rvi})]$ will be large. Unfortunately, this is not the case in general: Using properties of the $\invchi$ prior on $\rvSigma$, one can show that conditional on $\rvX$, the expectation of each term in $\cT_{\text{naive}}$ is $\frac{\rvq_j-1}{1+n\rvq_j}$, where $\rvq_j$ denotes the empirical variance in coordinate $j$. Thus, when $\rvq_j\gg 1$, the contribution of term $j$ is about $1/n$, but when $\rvq_j \in (0,1)$ the contribution is negative and is typically on the order of $-1/n$. It follows that while the top $k$-terms may contribute $k/n$ to the test, a careful selection algorithm may select many of the coordinates with the smallest variances, and these may contribute up to $-\ell/n$ to the test. Hence, this approach can only hope to show that $\ell$ must be at least $k$, which is too weak for our purposes (we aim to show that $\ell$ must be at least $d/\log d$ even when $k\approx n^2$). The problematic behavior in the test $\cT_{\text{naive}}$ occurs because the coordinates with the top-$k$ variances don't contribute enough to the sum to overcome the many small negative contributions from the coordinates where $\sigma^2_j\in(0,1)$. Thus, a natural way to circumvent this barrier is to scale each term by $\sigma_j$. Because the scaling is asymmetric, the result is that the large terms become much larger, while the impact on the small terms is negligible. This leads us to the following more effective test
\begin{align*}
    \cT_{s,\Sigma}(z) := \sum_{j \in [d]}\paren{\frac{z_j^2}{\sigma_j^2}-1}\min\left\{\sigma_j, \frac{d}{k}\right\}s_j,
\end{align*}
where the $\min$ is required for technical reasons (the $\sigma^2_j$'s are heavy tailed, and thus clipping helps to bound the variance). Using properties of the $\invchi(1,1)$ distribution, one can show that the top-$k$ variances are at least $(d/k)^2$, and thus, the contribution of the corresponding terms in $\cT$ increases by a factor of $d/k$ when compared to $\cT_{\text{naive}}$. Since the negative terms are essentially unaffected (they have small $\sigma_j$), we only require $\ell\leq d/\log d$ to guarantee that $\Ex[\cT(\rvx_{\rvi})]$ is large.

\subsection{Related work}

\paragraph{Differentially private Gaussian mean estimation.}
Differentially private mean estimation has been studied extensively, including asymptotically efficient estimators~\citep{Smith11} and estimators for univariate Gaussians with unknown parameters~\citep{KarwaV18}. A series of works established nearly optimal guarantees for Gaussian mean estimation with known covariance in {\em Mahalanobis} error metric~\citep{BunKSW19, KamathLSU19, AdenAliAK21, LiuKKO21}. \citet{BrownGSUZ21} developed optimal Gaussian mean estimators in Mahalanobis error metric with unknown covariance, followed by polynomial time algorithms matching these guarantees~\citep{BrownHS23,KuditipudiDH23}. 
As defined above, this loss is affine invariant and implies
\[
    \|\tilde\mu-\mu\|_2
    \leq
    d_\Sigma(\tilde\mu-\mu)\sqrt{\|\Sigma\|_{\mathrm{op}}}.
\]
However, it normalizes errors by the variance in each direction, including directions with very small variance, and therefore does not capture the full benefits of anisotropy under Euclidean error. 
Moreover, the sample size required by the methods in~\citep{BrownGSUZ21} is at least $n\gtrsim d/\eps$. Our complementary results show that a sample size of $n\gtrsim d$ is required to achieve any finite error in Mahalanobis distance, showing that this requirement is almost tight. 

For Euclidean error, the FriendlyCore framework~\citep{TsfadiaCKMS22} provides a filtering-based approach to private aggregation and underlies the general-purpose Gaussian mean estimator used as a benchmark in this work. Aiming to exploit covariance structure, \citet{AumullerLNP23} developed variance-aware private mean estimators for approximately-known diagonal covariance matrices, which achieve optimal error as long as $n\gtrsim \sqrt{d}/\eps$. \citet{DaganJYZZ24} established optimal dimension-free guarantees up to logarithmic factors when the covariance is known, and improved dimension dependence when it is diagonal and unknown. 
Under pure differential privacy, \citet{ChenDMM26} also obtain
covariance-dependent Euclidean guarantees for known-covariance
Gaussian mean estimation, in a regime requiring at least
$d/\eps$ samples.

Outside of differential privacy, ~\citep{MinasyanZ23} and ~\citep{depersin2022optimal} established optimal, dimension-free robust estimators for the mean of anisotropic Gaussian distributions under general norms.

\paragraph{Privately learning covariance structure.} 
Privately learning the covariance or a low-dimensional subspace can enable subsequent mean estimation with fewer samples or better accuracy, but these intermediate tasks can themselves require a lot of data. 

Lower bounds for private Gaussian covariance estimation were established by~\citep{KamathMS22} and strengthened by~\citep{narayanan25better, PortellaH25}. These results show that learning the covariance of a Gaussian distribution with constant spectral accuracy, which would suffice for subsequently using mean estimation algorithms with known covariance, requires $n\gtrsim d^{3/2}$ samples.  

\citet{SinghalS21} give dimension-independent sample bounds for exact recovery of a subspace under a general position assumption, as well as approximate recovery under a sufficiently large eigenvalue gap. \citet{Tsfadia24} studies subspace estimation in a distribution-free setting and further quantifies the role of eigenvalue gaps. 

Related barriers arise in private sparse covariance estimation and PCA, where row-column sparsity of the covariance matrix alone need not eliminate the polynomial dependence of the sample size on the ambient dimension~\citep{KumarPST26}. 
Our lower bounds establish the cost of unknown covariance directly for mean estimation, without requiring the estimator to recover the covariance or its principal directions.

\paragraph{Fingerprinting lower bounds.} 
Our lower bounds build on the intuitive connection between accurate estimation and membership inference. \citet{BunUV14} introduced fingerprinting codes as a tool for information-theoretic lower bounds in differential privacy, and~\citet{DworkSSUV15} developed robust tracing attacks for binary and Gaussian data. These techniques were further developed for Gaussian estimation by~\citet{KamathLSU19}, for exponential families by~\citet{KamathMS22}, and for general parametric models through score attacks by~\citet{CaiWZ23}. \citet{PeterTU24} develop padding-and-permuting fingerprinting codes that vary smoothly with the required accuracy. 

Our tracing arguments exploit priors over the covariance whose conditional distributions given the sample can be analyzed explicitly. 
The additional challenge in our setting is to exploit uncertainty about the covariance: we design joint priors over the mean and covariance, reduce mean estimation to covariance-dependent sampling or selection tasks, and construct tracing attacks for these tasks instead. 

\paragraph{Private selection.}
Our diagonal-covariance lower bound is particularly related to the top-$k$ selection problem of~\citet{SteinkeU17}, who extended fingerprinting arguments to selecting coordinates with large means. Related questions arise in private sparse mean estimation~\citep{CaiWZ19}. Our auxiliary task, called $(k,\ell)$-selection, instead asks for a list of length $\ell$ containing a constant fraction of the coordinates with the largest variances, and allows the list to contain a lot more than $k$ coordinates. A direct adaptation of the mean-selection tracing argument is not sufficient for our purposes: contributions from selected low-variance coordinates can cancel those from high-variance coordinates. We address this issue using a weighted, truncated test, adapted to our prior on the variances. 

\section{Lower bound for mean estimation with general covariance}
\label{sec:non_diag}

This section provides the lower bound of the Bayes risk for random non-diagonal covariance matrices, uniformly distributed among the orthogonal projectors on $k$-dimensional linear subspaces of $\R^d$. We define the prior $\cP_{\text{sub}}(d,k)$ over $(\rvmu,\rvSigma)$ as follows: Let 
$\cH_{d,k}$ be the uniform distribution 
over $d\times k$ matrices with orthonormal columns. Then
\begin{align}
    \label{eq:nd:prior}
    \paren{\rvmu,\rvSigma}\sim\cP_{\text{sub}}(d,k) \iff 
    \begin{cases}
        \rvmu\sim \cN(0,d^2 \id_d) \\\rvSigma = \rvD\rvD^\top \ \text{where}\ 
        \rvD\sim \cH_{d,k}.
    \end{cases}
\end{align}
This is equivalent to taking $\rvSigma$ to be the projector onto a uniformly-random (according to the Haar measure) $k$-dimensional subspace. We will omit the parameters $d$ and $k$ from the notation when they are clear from context. 

\begin{theorem}[Lower bound for private mean estimation with unknown covariance]
\label{thm:nd:estimation-lower-bound}
    There exists $c>0$ such that for all $d,k,n\in\N$ with $k\leq cd$ and $n\log n\leq ck$, every $(1,c/n)$-DP algorithm $\cA$ satisfies: \[
     \Pr_{\substack{(\bm{\mu},\bm{\Sigma})\sim\cP_{\text{sub}} \\ \rvX\sim\cN(\bm{\mu},\bm{\Sigma})^{\otimes n}\\ \text{coins of }\cA}}\brackets{\norm{\cA(\rvX) - \bm{\mu}}{2}\geq \frac{c\sqrt{dk}}{n}} \geq \frac 14.     
    \]   
    That is, for all $\eps\leq 1 $ and $\delta \leq \frac c n$, we have $\prisk\bparen{\cP_{\text{sub}},n} = \Omega\left(\frac{\sqrt{dk}}{n}\right)$.
\end{theorem}

Recall that the algorithm of \cite{DaganJYZZ24} for the known covariance case obtains error 
\[
\tilde O_{\eps,\delta}\paren{\sqrt{\frac{\tr\paren{\Sigma}}{n}} + \frac{\tr(\Sigma^{1/2})}{n}} = \tilde O_{\eps,\delta}\paren{\sqrt{\frac{k}{n}} + \frac{k}{n}}.
\]
Comparing the above to our lower bound for mean estimation with unknown covariance yields the following cost of not knowing the covariance in private mean estimation: In the setting of \Cref{thm:nd:estimation-lower-bound},
\begin{align}
    \label{eq:nd:known-unknown}
    \prisk(\cP_{\text{sub}},n) =\tilde\Omega\paren{\sqrt{\frac{d}{k}}} \cdot \pkrisk(\cP_{\text{sub}},n). 
\end{align}
We note that this blowup is tight, since it matches the error of the folklore algorithm (see \Cref{sec:folklore-estimator}). 
To be precise, for the family $\cP_{\text{sub}}$, and privacy parameters $\eps=1, \delta=c/n$, the folklore algorithm based on~\citep{TsfadiaCKMS22} and presented in~\Cref{sec:folklore-estimator} achieves, with constant probability, error
\[ O\left(
    \sqrt{\frac{k}{n}}
    +\frac{\sqrt{dk}\,(\log n)^{3/2}}{n}
  \right)\]
for $n$ larger than a universal constant. If $n\geq C k$ for some constant $C>0$, then one can learn the k-dimensional subspace of $\Sigma$~\citep[Theorem 1.1, $\ell=k-1$]{SinghalS21} and learn the mean with dimension-free error after projecting the data on this subspace. Our lower bound applies to the almost-complementary regime where $n\log n\leq ck$ and states that the dimension-dependent cost of privacy of the folklore algorithm is necessary up to polylogarithmic factors. 

To prove \Cref{thm:nd:estimation-lower-bound}, we first show how to transform any private and sufficiently accurate mean estimation algorithm into a private ``sampling'' algorithm that correlates with the population covariance matrix. 

\begin{definition}[$(\lambda,\alpha)$-sampler]\label{def:nd:sampler}Fix $d,k,n\in\N$ and $\alpha,\lambda>0$. An algorithm $\cM$ is a $(\lambda,\alpha)$-sampler for $\cP_{\text{sub}}$ (defined in \eqref{eq:nd:prior}) with sample complexity $n$ if \[
    \Pr_{\substack{\rvSigma\sim\cP_{\text{sub}} \\  
        \rvX\sim\cN(0,\rvSigma)^{\otimes n} \\ 
        \text{coins of }\cM}}
    \brackets{\cM(\rvX)^\top\rvSigma\cM(\rvX)
    \geq 
    \lambda^2}\ge \frac{1}{12} 
    ~~\text{and}~~ \norm{\cM(\rvX)}{2}\leq \alpha \, .
    \]
\end{definition}

\begin{remark}
    Not all values of $\lambda$ and $\alpha$ are compatible with the existence of a $(\lambda,\alpha)$-sampler. Trivially, it follows directly from the definition that necessarily $\alpha \ge \lambda$. In fact, we will show in \cref{subsec:nd:lower_bound_sampler} that, for $n\le \tilde O(k)$, the norm $\alpha$ must be significantly larger than $\lambda$.
\end{remark}

Our main technical contribution is a lower bound on the norm of any private $(\lambda,\alpha)$-sampler that uses few samples.

\begin{theorem}[Lower bound for private sampling]
    \label{thm:nd:sampler_lower_bound}
    There exists a constant $c>0$ such that for all $d,k,n\in\N$ and $\lambda>0$, if $k\leq cd$ and $n\log n\leq ck$ then every $(\lambda,\alpha)$-sampler for $\cP_{\text{sub}}$ with sample complexity $n$ that is $(1,c/n)$-DP requires $\alpha\geq c\lambda\sqrt{d/k}$.
\end{theorem}

To prove our main result, we show how to construct a private sampler from a private mean estimation algorithm.

\begin{lemma}[From mean estimation to sampling]
\label{lem:nd:sampler-reduction}
    There exists a constant $c>0$ such that for all $d,k,n\in\N$ with $n \le c k$ and $k\leq d$ the following holds: If $\cA$ is a $(1,c/n)$-DP algorithm satisfying
    $$
    \Pr_{\substack{(\rvmu,\rvSigma)\sim\cP_{\text{sub}} \\ \rvX\sim\cN(\rvmu,\rvSigma)^{\otimes n}\\ \text{coins of }\cA}}\left[\norm{\cA(\rvX)-\rvmu}{2}\leq \alpha \right]\geq \frac{3}{4},
    $$
    then there exists a $\paren{ck/n,\alpha}$-sampler for $\cP_{\text{sub}}$ with sample complexity $n$ that is $(1,c/n)$-DP.
\end{lemma}

We now combine \Cref{thm:nd:sampler_lower_bound,lem:nd:sampler-reduction} to complete the proof of the mean estimation lower bound.

\begin{proof}[Proof of \Cref{thm:nd:estimation-lower-bound}]
Let $c>0$ be a sufficiently small constant, and let $\cA$ be a $(1,c/n)$-DP algorithm achieving the accuracy guarantee in \Cref{lem:nd:sampler-reduction} with parameter $\alpha$. Then there exists a $(1,c/n)$-DP mechanism $\cM$ that is a $(ck/n,\alpha)$-sampler for $\cP_{\text{sub}}$ with sample complexity $n$. It follows from \Cref{thm:nd:sampler_lower_bound} that for $n\log n<ck$, we must have $\alpha \geq \Omega(\sqrt{dk}/n)$, which yields the result.  
\end{proof}

The remainder of the section is organized as follows: In \Cref{subsec:nd:reduction} we prove \Cref{lem:nd:sampler-reduction} (the reduction from private mean estimation to private sampling), and in \Cref{subsec:nd:lower_bound_sampler} we prove \Cref{thm:nd:sampler_lower_bound} (the lower bound on the norm of the sampler). 

\subsection{Reduction from sampling}
\label{subsec:nd:reduction}

In this section, we construct a sampler that uses a mean estimation algorithm as a subroutine. The sampler used in our lower bound consists essentially of the clipped error of the mean estimation mechanism. 

\begin{proof}[Proof of \Cref{lem:nd:sampler-reduction}]    For a vector $y\in\R^d$ we define $\mathrm{proj}_\alpha(y)$ as the projection of $y$ into the $\ell_2$ ball of radius $\alpha$ centered at $0$. For $\rho =d$, let $\cM$ denote the following algorithm: 
    \begin{align}
     \text{$\cM(\rvX):$ Sample $\rvmu \sim \mathcal{N}\left(0,\rho^2 \id_d\right)$. Return $\mathrm{proj}_{\alpha}(\mathcal{A}(\rvX+\rvmu) -\rvmu)$.}    
    \end{align}

Since $\cA$ is $(\eps,\delta)$-DP, postprocessing immediately implies that $\cM$ is $(\eps,\delta)$-DP. It remains to show that $\cM$ is a $(\lambda,\alpha)$-sampler for appropriate $\lambda$. Recall that for every $\Sigma$ we denote $\norm{\cdot}{\Sigma}$ the pseudo-norm induced by $\Sigma$, e.g.  $\norm{\cM(\rvX)}{\Sigma} = \sqrt{\cM(\rvX)^\top\Sigma\cM(\rvX)}$. Let $G_1 := \{\norm{\cA(\rvX+\rvmu)-\rvmu}{2} \le \alpha\}$ be the event under which the clipping does not modify anything, and for sufficiently small $c>0$, let  $G_2 := \set{\|\cA(\rvX+\rvmu) - \rvmu\|_{\rvSigma}\ge \frac{ck}{n}}$ be the event that the raw output is large. Then, conditioned on $G_1$ and $G_2$ we have that
    \begin{align*}
        \norm{\proj_{\alpha}(\cA(\rvX+\rvmu)-\rvmu)}{\rvSigma} = \norm{\cA(\rvX+\rvmu) - \rvmu}{\rvSigma}\geq \frac{ck}{n}.
    \end{align*}
    We show that both events hold with probability $1/12$. Recall that by the guarantee of $\cA$ we have $\Pr[\norm{\cA(\rvX+\bm{\mu}) - \bm{\mu}}{2}\leq \alpha]\geq 3/4$, and thus $\Pr[G_1]\geq 3/4$. Next we show that $\Pr[G_2]\geq 1/3$. Our proof uses the following folklore result, which is a Bayesian variation of well-known lower bounds on private mean estimation under a Gaussian distribution via fingerprinting arguments \citep{KamathLSU19}, the specific statement we use is from \cite{GaucherLSUZ26}.
    
    \begin{proposition}[Folklore Result]
    \label{prop:nd:folklore-dp-lb}
    Let $k,n \in \N$. There exist absolute constants $c,C>0$, such that for any $(\eps,\delta)$-mechanism $\cA:\R^{k\times n}\rightarrow \R^k$ with $\delta \le \frac{c}{n}$, any $\eps \le 1$, and any $\rho \ge C\max\paren{1,\sqrt{k}/n}$,
    \[
    \Pr_{\substack{\rvmu \sim \cN(0,\rho^2\id_k) \\ \rvX \sim \cN(\rvmu,\id_k)^{\otimes n}\\ \text{coins of }\cA}}\brackets{\norm{\cA(\rvX) - \rvmu}{2} \ge \frac{c k}{n}} \ge \frac{1}{3}.
    \]
    \end{proposition}

    Fix $\Sigma=DD^\top$. We show that there exists an algorithm $\cA_{\Sigma}$ such that if $\rvmu_k\sim\cN(0,\rho^2\cdot \id_k)$ and $\rvY\sim\cN(\rvmu_k, \id_k)^{\otimes n}$ then $\norm{\cA_{\Sigma}(\rvY) - \rvmu_k}{2}=\norm{\cA(\rvX) - \rvmu}{\Sigma}$, where $\rvX\sim\cN(\rvmu,\Sigma)^{\otimes n}$ and $\rvmu\sim\cN(0,\rho^2\cdot \id_d)$. This will allow us to use \Cref{prop:nd:folklore-dp-lb} to bound the probability of event $G_2$. Let $\cA_{\Sigma}$ denote the following algorithm,
    \begin{align}
    \text{$\cA_\Sigma(\rvY)$ ~:~ Sample $\rvmu_{\bot}\sim\cN(0,\rho^2\cdot(\id_d-\Sigma))$. Set $\rvx_i=D \rvy_i + \rvmu_{\bot}$. Return $D^\top\cA(\rvX)$.} 
    \end{align}
    Using the fact that $D^\top D=\id_k$ and $D^\top\rvmu_{\bot}=0$ we see that
    \begin{align*}
        \norm{D^\top\cA(\rvX)-\rvmu_k}{2} 
        &= \norm{D^\top\cA(\rvX)-D^\top D\rvmu_k}{2}\\
        &= \norm{D^\top(\cA(\rvX)- D\rvmu_k -\rvmu_{\bot})}{2}\\
        &= \norm{\cA(\rvX)-\rvmu}{\Sigma} 
    \end{align*}
    
    By the distribution of $\rvmu_k$ and $\rvmu_{\bot}$, the distribution of $\rvmu = D\rvmu_k + \rvmu_{\bot}$ is $\cN(0,\rho^2\cdot\id_d)$. Moreover, by the distribution of $\rvy_i$, we have, conditioned on $\rvmu$, that $\rvx_i\sim\cN(\rvmu,\Sigma)$. Thus, the event $G_2$ defined above is equivalent to the event that $\norm{\cA_{\rvSigma}(\rvY)-\rvmu_k}{2}\geq \frac{ck}{n}$, which, invoking \Cref{prop:nd:folklore-dp-lb} as $\rho = d \ge k\ge c^{-\nicehalf}\max(1,\sqrt{k}/n)$, occurs with probability at least $\frac 13$.

    A union bound over $G_1$ and $G_2$ now yields that $\Pr[\norm{\cM(\rvX)}{\bm{\Sigma}}\geq ck/n]\geq 1/12$. Moreover, the definition of $\cM$ enforces that $\norm{\cM(\rvX)}{2}\leq \alpha$, and thus $\cM$ is a $(ck/n,\alpha)$-sampler.
\end{proof}

\subsection{Private sampling lower bound}\label{subsec:nd:lower_bound_sampler}

We consider covariance $\rvSigma\sim\cP_{\text{sub}}$, dataset $\rvX=\rvx_1,\dots,\rvx_n\sim\cN(0,\rvSigma)^{\otimes n}$, output $\rvy=\cM(\rvX)$, test point $\rvz\sim\cN(0,\rvSigma)$. To prove a lower bound on the norm of the sampler, we consider the following test statistic. For fixed $\Sigma$ and $y,z\in\R^d$ with $y^\top\Sigma y>0$ let 
\[
T_{\Sigma,y}(z) = \frac{\ip{y}{z}^2}{y^\top\Sigma y}.
\]
If $y^\top\Sigma y=0$, then we set $T_{\Sigma,y}(z)=0$. We will omit the dependence on $\Sigma$ and $y$ when they are clear from context and instead write $T(z)$. \Cref{thm:nd:sampler_lower_bound} follows immediately from the following two Lemmata: Our first result provides a bound on the test in the OUT case that holds for all fixed $y$ and $\Sigma$ with high probability over the test point $\rvz$. 

\begin{lemma}[OUT case upper bound]
    \label{lem:nd:out-case-bound} For every $c\in(0,1)$ there exists $C>0$ such that for all $y \in \R^d$ and $\Sigma \succeq 0$, if $k\geq Cn\log n$ then 
    \[
    \Pr_{\rvz \sim\cN(0,\Sigma)}\brackets{T_{\Sigma,y}(\rvz)\geq \frac {k}{4n}}\leq \frac{c}{n}.
    \]
\end{lemma}

\begin{proof}[Proof of \Cref{lem:nd:out-case-bound}]
    The proof follows from the distribution of $T_{\Sigma,y}(\rvz)$ for fixed $y$ and $\Sigma$. Observe that $\ip{y}{\rvz}\sim\cN(0,y^\top\Sigma y)$, and thus $\ip{y}{\rvz} / \sqrt{y^\top\Sigma y}\sim\cN(0,1)$. Since $T_{\Sigma,y}(\rvz) = (\ip{y}{\rvz} / \sqrt{y^\top\Sigma y})^2$ we see that $T_{\Sigma,y}(\rvz)\sim\chi^2_{1}$. Standard concentration bounds for chi-square random variables (\Cref{lem:laurent-massart}) yield the result. 
\end{proof}

Next, we state an analogous result for the value of the test in the IN case. We show that over the randomness of the data $\rvX$, the mechanism's output $\rvy=\cM(\rvX)$, the covariance $\rvSigma$, and the choice of index $\rvi$, $T\paren{\rvx_{\rvi}}$ is ``large'' with probability $\Omega(1/n)$. 

\begin{lemma}[IN case lower bound]
    \label{lem:nd:in-case-bound} Let $c>0$ be a sufficiently small constant. Suppose $n\leq ck$ and $\alpha\leq c\lambda\sqrt{d/k}$. Then, if $~\rvi\sim[n]$ is chosen uniformly 
    \[
    \Pr_{\rvX,\rvy,\rvSigma,\rvi}\brackets{T_{\rvSigma,\rvy}(\rvx_{\rvi})\geq \frac {k}{4n}}\geq \frac cn.
    \]
\end{lemma}

\begin{proof}[Proof of \Cref{lem:nd:in-case-bound}]
Throughout the proof, we let $C>0$ denote a sufficiently large constant and $c>0$ a small constant. We show that there exists an event $G$ over the sampling of $\rvX,\rvy$ and $\rvSigma$ such that $\Pr[G]\geq c$, and for all $(X,y,\Sigma)$ in $G$, the expectation of $T(\rvx_{\rvi})$ over $\rvi$ is at least $4k/n$. Thus, there is some $i$ such that $T(\rvx_i)\geq 4k/n$, which, since $\rvi$ is uniform in $[n]$ will suffice to prove the lemma. We begin by defining the event $G$. 

Let $P_X$ be the orthogonal projection onto the span of $X$, and let $\hat\Sigma_X=\frac{1}{n}XX^\top$. Define $G = G_1\cap G_2 \cap G_3$ as the event that the following three conditions hold:
\begin{enumerate}
    \item[$G_1$:] $\frac{k}{2n}\cdot P_X \preceq \hat \Sigma_X\preceq \frac{2k}{n}\cdot P_X$.
    \item[$G_2$:] $y^\top(\Sigma - P_X)y\leq \frac{2C\alpha^2k}{d}$.
    \item[$G_3$:] $y^\top\Sigma y\geq \lambda^2$.
\end{enumerate}

Intuitively, the event $G$ occurs if $X$ is a typical sample, if the norm of $y$ in the $\Span(\Sigma)\setminus\Span(X)$ is at most a $\frac{k}{d}$ fraction of the total norm of $y$, and if the norm of $y$ in $\Span(\Sigma)$ is at least the required norm $\lambda^2$. We first show that the expectation over $\rvi$ is large for every $X,y$ and $\Sigma$ in $G$.
\begin{claim}
    \label{claim:nd:expectation-over-I} Suppose that $\alpha\leq \lambda\sqrt{\frac{d}{4Ck}}$ and that $X,y$, and $\Sigma$ are in $G$. Then 
    $\Ex_{\rvi}\brackets{T_{\Sigma,y}(x_{\rvi})}\geq \frac {k}{4n}.
    $
\end{claim}

\begin{proof}
    The essence of the proof is the observation that 
    $$\Ex_{\rvi}[T(x_{\rvi})]=\frac{y^\top\Ex[x_{\rvi}{x_{\rvi}}^{\!\top}]y}{y^\top\Sigma y}=\frac{y^\top\hat\Sigma_Xy}{y^\top\Sigma y}.$$
    The remainder of the proof follows by algebraic manipulation, and by applying the inequalities that hold for all $X,y$, and $\Sigma$ in $G$. 

    \begin{align*}
        \frac{y^\top\hat\Sigma_Xy}{y^\top\Sigma y}
        &\geq \frac{k}{2n}\cdot \frac{y^\top P_Xy}{y^\top\Sigma y} \tag{$G_1$}\\
        &\geq \frac{k}{2n}\cdot\frac{y^\top\Sigma y - 2C\alpha^2k/d}{y^\top\Sigma y} \tag{$G_2$}\\
        &= \frac{k}{2n}\paren{1-\frac{2C\alpha^2k}{dy^\top\Sigma y}}
        \geq \frac{k}{2n}\paren{1-\frac{2C\alpha^2k}{d\lambda^2}} \tag{$G_3$}\\
        &\geq \frac{k}{4n}.\qedhere
    \end{align*}
    
\end{proof}

Our second claim states that $G$ occurs with constant probability.

\begin{claim}
    \label{claim:nd:good-event-probability}
    There exists a constant $c>0$ such that for all sufficiently large $k\leq cd$ and $n\leq ck$, we have $\Pr_{\rvX,\rvy,\rvSigma}[G]\geq c$.
\end{claim}

\begin{proof}
    Let $C>0$ be a sufficiently large constant. And recall that $G_1$, $G_2$ and $G_3$ are the events composing $G$. We first bound $G_1$ and $G_3$: By standard results in random matrix theory (see \Cref{lem:random-matrix-concentration}), as long as $k$ is at least some sufficiently large constant, $\Pr[G_1]\geq 1-1/C$. Additionally, by the definition of an $(\lambda,\alpha)$-sampler (\Cref{def:nd:sampler}) we have $\Pr[G_3]\geq \frac{1}{12}$.  

    It remains to bound the probability of $G_2$. The key idea is that since $\rvSigma$ is uniformly distributed on the space of rank $k$ orthogonal projections in $\R^d$, conditioned on $\rvX$, the matrix $\rvSigma-P_{\rvX}$ is a rank $k-n$ random orthogonal projection within the linear subspace $\Span(P_{\rvX})^{\perp}$, which is a $d-n$ dimensional space. Thus,
    $$\Ex[\rvSigma-P_{\rvX}\mid \rvX] = \frac{k-n}{d-n}(\id_d - P_{\rvX})\preceq \frac{2k}{d}\id_d.$$
    Since $\norm{\rvy}{2}\leq \alpha$ we have that $\Ex[\rvy^\top(\rvSigma-P_{\rvX})\rvy]\leq 2\alpha^2k/d$, and hence, by Markov's inequality, 
    $$\Pr[G_2]\geq 1-\Pr\big[\rvy^\top(\rvSigma-P_{\rvX})\rvy\geq C\Ex[\rvy^\top(\rvSigma-P_{\rvX})\rvy]\big]\geq 1-\frac{1}{C}.$$

    To complete the proof of the claim, observe that the union bound yields $\Pr[G_1\cap G_2\cap G_3]\geq \frac{1}{12} - \frac{2}{C}\geq c$ for $C\geq 48$ and $c\leq 1/24$. 
\end{proof}

We now complete the proof of \Cref{lem:nd:in-case-bound}. Since for every $X,y,\Sigma$ in $G$, there exists $i^*$ such that $T(\rvx_{i^*})\geq \Ex_{\rvi}[T(\rvx_{\rvi})]$, we can conclude that
\begin{align*}
    \Pr_{\rvX,\rvy,\rvSigma,\rvi}\brackets{T(\rvx_{\rvi})\geq \frac{k}{4n}}
    &\geq \Pr\brackets{T(\rvx_{\rvi})\geq \frac{k}{4n}\mid G}\Pr[G]\\
    &\geq \Pr\brackets{T(\rvx_{\rvi})\geq \Ex_{\rvi}\brackets{T(\rvx_{\rvi})\mid \rvX,\rvy,\rvSigma}\mid G}\Pr[G]\\
    &\geq \Pr[\rvi=i^*]\Pr[G]\\
    &\geq \frac{c}{n}.
\end{align*}
The second inequality follows from \Cref{claim:nd:expectation-over-I} and the last inequality follows from \Cref{claim:nd:good-event-probability} and the fact that $\rvi$ is uniform in $[n]$ and independent of $\rvX,\rvy$ and $\rvSigma$. 
\end{proof}

We conclude this section by combining \Cref{lem:nd:in-case-bound,lem:nd:out-case-bound} to prove \Cref{thm:nd:sampler_lower_bound}, which states that every $(\lambda,\alpha)$-sampler that is $(1,c/n)$-DP and uses $n\leq k/\log n$ samples has norm $\alpha\geq c\lambda\sqrt{d/k}$.  It will be convenient to define the following notation for indistinguishability.

\begin{definition}[$(\eps,\delta)$-indistinguishability] 
\label{def:indistinguishability}
Two distributions $\cP,\cQ$ over a domain $\cW$ are $(\eps,\delta)$-indistinguishable, denoted $\cP \approx_{\eps,\delta} \cQ$, if, for any measurable subset $W \subset \cW$,
\[
\cP(W) \le e^{\eps}\cQ(W) + \delta \quad \text{ and } \quad \cQ(W) \le e^{\eps}\cP(W) + \delta.
\]
If $\rvx$ and $\rvx'$ are random variables with distribution $\cP$ and $\cQ$ respectively, we write  $\rvx \approx_{\eps,\delta}\rvx'$ to indicate that $\cP \approx_{\eps,\delta} \cQ$. 
\end{definition}

\begin{proof}[Proof of \Cref{thm:nd:sampler_lower_bound}]
The proof follows by contradiction. Suppose $\cM$ is an $(\eps,\delta)$-DP $(\lambda,\alpha)$-sampler with $\eps\leq 1$, $\delta\leq c/2n$ and $\alpha\leq c\lambda\sqrt{d/k}$. For every index $i\in [n]$ let $\rvX^{\oplus i}$ denote $\rvX$ with $\rvx_i$ replaced by $\rvz$. Then, since $\cM$ is $(\eps,\delta)$-DP we have 
\[
T_{\cM(\rvX)}(\rvz)\approx_{\eps,\delta} T_{\cM(\rvX^{\oplus i})}(\rvz)\equiv T_{\cM(\rvX)}(\rvx_i)
\]
where the equality in distribution follows from the fact that $\rvx_i$ and $\rvz$ are both sampled from $\cN(0,\Sigma)$ independently of the other datapoints. Hence, if $\rvy=\cM(\rvX)$ and $\rvi\sim[n]$, then by \Cref{lem:nd:in-case-bound,lem:nd:out-case-bound} there exists a constant $C>0$ (depending only on $c$) such that if $k\geq Cn\log n$
\[
\frac cn\leq \Pr\brackets{T_{\cM(\rvX)}(\rvx_{\rvi})\geq \frac{k}{4n}}\leq e^{\eps}\Pr\brackets{T_{\cM(\rvX)}(\rvz)\geq \frac{k}{4n}} + \delta < \frac{c}{2n} + \frac{c}{2n},
\]
which is a contradiction. Therefore, every such sampler must have $\alpha\geq c\lambda\sqrt{d/k}$.
\end{proof}

\section{Lower bound for mean estimation with diagonal covariance}
\label{sec:diagonal}

In this section we show a lower bound for private mean estimation when the covariance matrix is promised to be diagonal. Our lower bound uses the scaled-inverse-chi-squared distribution as a prior over the variances in each coordinate. 
\begin{definition}[$\invchi(\tau,\nu)$]
     \label{def:inv-chi}
The scaled-inverse-chi-squared distribution with $\nu$ degrees of freedom and scale $\tau$ is the law with density, for $x \ge 0$, given by
\begin{align*}
    f_{\tau,\nu}(x)\propto \frac{\exp\brackets{\frac{-\nu \tau^2}{2x}}}{x^{1+\nu/2}}.
\end{align*}
Equivalently $\sigma^2 \sim \invchi(\nu,\tau^2)$, if $\displaystyle \frac{\nu \tau^2}{\sigma^2}\sim \chi^2_{\nu}$.
\end{definition} 
We then define the prior $\cP_{\text{diag}}$ over instance $(\rvmu,\rvSigma)$ as follows: 
\begin{align}
    \label{prior:diagonal}
    (\rvmu,\rvSigma) \sim \cP_{\text{diag}}(d) \iff 
    \begin{cases}
        \rvmu \sim \cN(0, d^2\id_d)\\
        \rvSigma = \mathrm{Diag}((\rvsigma^2_i)_{i\in[d]}), ~ \text{where} ~   (\rvsigma^2_i)_{i\in[d]} \sim \invchi(1,1)^{\otimes d}
    \end{cases}
\end{align}
with $\rvmu$ and $\rvSigma$ independent. We now state our lower bound for private mean estimation with unknown diagonal covariance. 

\begin{theorem}
\label{thm:diag:estimation-lb}
     There exists $c>0$ such that for all $d,n\in\N$ with $n\log n\leq c\sqrt{d}$, every $(1,c/n)$-DP algorithm $\cA$ satisfies
    \[
    \Pr_{\substack{\rvX \sim \cN(\rvmu,\rvSigma)^{\otimes n}\\ (\rvmu,\rvSigma)\sim \cP_{\text{diag}} \\ \text{coins of $\cA$}}}\brackets{\|\cA(\rvX) - \rvmu\|_2 \geq \frac{cd^{3/2}}{n^2\log d}} \ge \frac{1}{4}.
    \]
    That is, for all $\eps\leq 1$ and $\delta\leq \frac{c}{n}$ we have $\prisk\paren{\cP_{\text{diag}},n}=\tilde\Omega\paren{\frac{d^{3/2}}{n^2}}$.
\end{theorem}
Recall that on instances $(\rvmu,\rvSigma)\sim\cP_{\text{diag}}$ the known-covariance algorithm of \cite{DaganJYZZ24} obtains error 
$$
\tilde O_{\eps,\delta}\paren{\frac{d}{\sqrt{n}}},
$$
since $\tr(\Sigma)=\Theta(d^2)$ and $\tr(\Sigma^{1/2})=\Theta(d\log d)$ with high probability. Thus we obtain the following scale-free ``cost'' of not knowing the covariance in private mean estimation:
\begin{align}
    \label{eq:diag:known-unknown-comparison}
    \prisk\paren{\cP_{\text{diag}},n} = \tilde\Omega\paren{1+\frac{\sqrt{d}}{n^{3/2}}}\cdot \pkrisk\paren{\cP_{\text{diag}},n}.
\end{align}

We note that the blow-up in error expressed by \eqref{eq:diag:known-unknown-comparison}, matches the error of the unknown covariance algorithm from \cite{DaganJYZZ24}. To be precise, for the family $\cP_{\text{diag}}$ and parameters $\eps=1, \delta=c/n$, the method in~\citep[Theorem 4.1]{DaganJYZZ24} achieves, with constant probability, error 
\[O\Bigg(\frac{d}{\sqrt n}
    +\frac{d\log d\,(\log n)^{3/2}}{n}\\[-2pt]
    +\frac{d^{3/2}(\log d)^2(\log n)^2\log(dn)}{n^2}
  \Bigg)\]
for $n\geq C\sqrt{\log n}\,\log d\,\log(dn)$ for universal constant $C>0$.

When the sample-size is large enough so that $n\log n \geq \sqrt{d}$, the middle term dominates the cost of privacy, and it is tight up to polylogarithmic factors, by the known-covariance lower bound~\citep{DaganJYZZ24}. The remaining sample-size regime of $n\log n\leq \sqrt{d}$, is exactly the regime where the third term dominates and matches the regime of our new lower bound, showing that this term is also indeed necessary, up to polylogarithmic factors. 

The proof of \Cref{thm:diag:estimation-lb} is similar to  \Cref{thm:nd:estimation-lower-bound}, but instead of reducing from a sampling problem, we reduce from the following selection problem:

\begin{definition}[$(k,\ell)$-selection mechanism]
\label{def:selection_mechanism}
Let $d, n,k,\ell \in \N$. An algorithm $\cM$ is a $(k,\ell)$-selection mechanism for $\cP_{\text{diag}}$ with sample complexity $n$ if the output $\rvs = \cM(\rvX) \in \set{0,1}^{d}$ of the algorithm satisfies:
    \begin{enumerate}
        \item \textbf{Boundedness:} \(\sum_{j=1}^d \rvs_j \le \ell\).
        \item \textbf{Accuracy:} Let $\bm{\pi}$ be the permutation such that $\rvsigma^2_{\bm{\pi}(1)}\ge\ldots\ge\rvsigma^2_{\bm{\pi}(d)}$. Then \[
        \Pr_{\substack{\rvSigma\sim\cP_{\text{diag}} \\ \rvX\sim\cN(0,\rvSigma)^{\otimes n} \\ \text{coins of $\cM$}}}\left[\sum_{j=1}^k \rvs_{\bm{\pi}(j)} \ge \frac{k}{4}\right]\geq \frac{1}{3}.
        \]
    \end{enumerate}  
    In other words, a selection mechanism outputs no more than $\ell$ indices, and with probability $1/3$ over the randomness of the data and the algorithm, it selects at least $k/4$ indices from the indices with the $k$ largest variances.
\end{definition}

The next theorem shows a tradeoff between the accuracy and the boundedness of private $(k,\ell)$-selection mechanisms.

\begin{theorem}[Lower bound for private selection]
    \label{thm:diag:selection-lb}
    There exist constants $C,c>0$ such that for all $d,k,n\in\N$ with $C\max\set{n^2\log n, \log d}\leq k\leq cd$, every $(k,\ell)$-selection mechanism for $\cP_{\text{diag}}$ with sample complexity $n$ that is $(1,c/n)$-DP must have  
    $ \ell\geq d/C\log(d).
    $
\end{theorem}

Informally, \Cref{thm:diag:selection-lb} states that a private $(k,\ell)$-selection mechanism which successfully selects a constant fraction of the $n^2\log n$ largest variance coordinates, must at the same time select almost all the $d$ coordinates overall. The proof is deferred to \Cref{sec:diag:selection-lb-proof}. Next, we state \Cref{lem:diag:selection-reduction}, the reduction from private selection, and complete the proof of our main result (\Cref{thm:diag:estimation-lb}).

\begin{lemma}[From mean estimation to selection]
\label{lem:diag:selection-reduction}
    There exist constants $C,c>0$ such that for all $\alpha>0$, and $n,d,k\in \N$ with $Cn\leq k\leq d$, the following holds: If $\cA$ is a $(1,c/n)$-DP algorithm satisfying
    $$
    \Pr_{\substack{(\rvmu,\rvSigma)\sim\cP_{\text{diag}} \\ \rvX\sim\cN(\rvmu,\rvSigma)^{\otimes n} \\ \text{coins of $\cA$}}}\left[\norm{\cA(\rvX)-\rvmu}{2}\leq \alpha\right]\geq \frac{3}{4},
    $$
    then, for $\ell=\lceil\frac{C\alpha^2 kn^2}{d^2}\rceil$, there exists a $(1,c/n)$-DP mechanism $\cM$ that is a $(k,\ell)$-selection mechanism for $\cP_{\text{diag}}$ with sample complexity $n$. 
\end{lemma}

We now combine \Cref{lem:diag:selection-reduction} and \Cref{thm:diag:selection-lb} to prove the mean estimation lower bound. 

\begin{proof}[Proof of \Cref{thm:diag:estimation-lb}]
\label{proof:thm:diag:estimation-lb}
    Let $c,C>0$ and $\eps,\delta\in(0,1)$ be as in \Cref{thm:diag:selection-lb}. Fix $n\in\N$ and set $k=C\max\set{n^2\log n, \log d}$. Suppose $\cA$ is an $(\eps,\delta)$-DP algorithm with sample complexity $n$, that achieves error $\alpha$ with probability at least $2/3$ on instances sampled from prior $\cP_{\text{diag}}$. Then, by \Cref{lem:diag:selection-reduction}, there exists an $(\eps,\delta)$-DP $(k,\ell)$-selection mechanism with $\ell\leq C\alpha^2kn^2/d^2$. But by our setting of $k$, \Cref{thm:diag:selection-lb} implies that $\ell\geq d/C\log d$, and therefore $\alpha^2\geq c^2d^3/Ckn^2\log d\geq \Omega(d^3/n^4\log(d)^2)$.    
\end{proof}

\subsection{Reduction from selection}

In this section, we complete the proof of \Cref{lem:diag:selection-reduction} by using a private mean estimation algorithm to design a $(k,\ell)$-selection mechanism. 
Intuitively, the selection mechanism selects the coordinates in which the mean estimator suffers error above a carefully chosen threshold, and among these, we expect to find a lot of the coordinates with large variance. 

\begin{proof}[Proof of \Cref{lem:diag:selection-reduction}]
Let $\cA$ be the algorithm with accuracy $\alpha$ as in the statement of \Cref{lem:diag:selection-reduction}. Define the selection mechanism $\cM$ with fixed parameters $\tau=\frac{d}{Cn\sqrt{k}}$ and $\rho=d$ as follows:
\begin{align}
\label{alg:diag:selection-reduction}
\begin{aligned}
        \cM(\rvX) :~ 
        &\text{Sample $\rvmu \sim \cN(0,\rho^2\id_d)$ and set $\rvY=\rvX+\rvmu$.} \\
        &\text{For $j\in[d]$ set $\rvs_j = \Ind\left[\left|\cA(\rvY)_j-\rvmu_j\right|\ge \tau\right]$.} \\
        &\text{If $\sum \rvs_j\leq \ell$ return $\rvs=(\rvs_1,\dots,\rvs_d)$, and otherwise return $\bm{0}$.}
\end{aligned}
\end{align}
To show that $\cM$ is a $(k,\ell)$-selection mechanism, we must show that it never selects more than $\ell$ coordinates, and that with probability at least $1/4$ it selects at least $k/4$ of the coordinates with the $k$ largest variances. By inspection of \eqref{alg:diag:selection-reduction}, one can easily see that $\cM$ never outputs a selection of more than $\ell$ coordinates.

Let $\rvJ=\{\bm{\pi}(1), \dots, \bm{\pi}(k)\}$ be the coordinates of the largest $k$ variances in $\rvSigma$. In the remainder of the proof, we argue that $\cM$ outputs at least $k/4$ of the coordinates in $\rvJ$ with probability at least $1/4$. First we argue that $\cM$ typically does not return $\bm{0}$: By construction of $\cM$, 
for any choice of $\tau$, we have $\tau^2\sum_{j=1}^d \rvs_j \leq \|\cA(\rvY)-\rvmu\|^2_2$, and therefore, setting $\ell=\lceil \alpha^2/\tau^2\rceil$ and using the accuracy guarantee of $\cA$,  
\begin{align}
\label{eq:boundedness-prob}
\Pr_{\substack{(\rvmu,\rvSigma)\sim \cP_{\text{diag}} \\ \rvX \sim \cN(0,\rvSigma)^{\otimes n}}}\brackets{\sum_{j=1}^d \rvs_j\leq \ell}
\geq \Pr_{\substack{(\rvmu,\rvSigma)\sim \cP_{\text{diag}} \\ \rvY\sim\cN(\rvmu,\rvSigma)^{\otimes n}}}\brackets{\|\cA(\rvY) - \rvmu\|\le \alpha}\ge \frac{3}{4}.
\end{align}

Next, we argue that $\rvs_j=1$ for at least $k/4$ of the indices in $\rvJ$ with probability at least $1/4$. Our proof leverages a version of the standard lower bound for private mean estimation that bounds the number of coordinates with error above some threshold. 

\begin{proposition}[Coordinate-wise error lower bound \cite{GaucherLSUZ26}]
\label{prop:coordinate-wise-error}
There exist constants $C,c>0$ such that for all $n,d\in\N$ and
$\alpha,\rho>0$ satisfying
\[
    \rho\geq C\max\{1,\alpha\},
    \qquad
    d\geq C(n+\alpha^2n^2),
\]
if $\rvmu\sim \cN(0,\rho^2\id_d)$ and $\rvX=\rvx_1,\dots,\rvx_n\sim\cN(\rvmu,\id_d)^{\otimes n}$, then every $(1,c/n)$-DP mechanism $\cA$ satisfies 
\[
    \Pr_{\substack{\rvmu\sim\cN(0,\rho^2\id_d) \\ \rvX\sim\cN(\rvmu,\id_d)}}
    \left[
        \abs{\{j\in[d] : |\cA(\rvX)_j-\rvmu_j|\geq\alpha\}}\geq d/2
    \right]\geq\frac23.
\]
\end{proposition}

Since \Cref{prop:coordinate-wise-error} is for identity covariance, a few extra steps will be needed before we can apply it to our setting. Informally, we will construct an algorithm $\cA_{\rvSigma}$ that uses $\cA$ as a subroutine for a $k$-dimensional mean estimation problem with identity covariance. We then show that the per-coordinate error of $\cA_{\rvSigma}$ can be bounded by the error of $\cA$ on the coordinates in $\rvJ$. Applying \Cref{prop:coordinate-wise-error} to $\cA_{\rvSigma}$ allows us to conclude that $\cA$ must have large error in many of the coordinates in $\rvJ$.  

Consider the following $k$-dimensional mean estimation problem:
\begin{align}
    \label{eq:k-dim-experiment}
    \rvtheta\sim \cN\left(0, \frac{4k^2\rho^2}{d^2}\id_k\right) ~~ \text{and} ~~ \rvZ=\rvz_1,\dots,\rvz_n \sim \cN\left(\rvtheta, \id_k\right)^{\otimes n}.
\end{align}

For a vector $y\in\R^d$ and $J\subseteq[d]$, let $y_J$ denote the vector $(y_j)_{j\in J}$, and similarly, for a matrix $\Sigma\in\R^{d\times d}$, let $\Sigma_J$ denote the $|J|\times |J|$ matrix with entries $\Sigma_{j,j'}$ for $j,j'\in J$. Define algorithm $\cA_{\rvSigma}$ as follows: 
\begin{align}
    \begin{aligned}
        \cA_{\rvSigma}(\rvZ) :~ &\text{If $\rvsigma^2_{\rvpi(k)}<\frac{d^2}{4k^2}$ return $\bot$. Otherwise continue.}\\
        &\text{Sample $\rvtheta'_{\overline{\rvJ}}\sim\cN(0,\rho^2\id_{d-k})$ and for each $i\in[n]$:}\\
        &\text{\qquad Sample $\tilde\rvy_{i,\rvJ}=\frac{d}{2k}\rvz_{i} + \rvw_{i}$ where $\rvw_{i}\sim\cN\left(0,\rvSigma_{\rvJ}-\frac{d^2}{4k^2}\id_k\right)$.}\\
        &\text{\qquad Sample $\tilde\rvy_{i,\overline{\rvJ}}\sim \cN\paren{\rvtheta'_{\overline{\rvJ}}, \rvSigma_{\overline{\rvJ}}}$}\\
        &\text{Return $\frac{2k}{d}\cA(\tilde{\rvY})_{\rvJ}$}
    \end{aligned}
\end{align}

To see that $\cA_{\rvSigma}$ is private, observe that since $\cA$ is $(\eps,\delta)$-DP, and for fixed random coins, $\cA_{\rvSigma}$ simply runs $\cA$ on an affine transformation of the data and outputs the result, we have that $\cA_{\rvSigma}$ is also $(\eps,\delta)$-DP.

Next, we relate the accuracy of $\cA_{\rvSigma}$ on the task defined in \eqref{eq:k-dim-experiment}, to the accuracy of $\cA$ on the coordinates in $\rvJ$ in the experiment in the statement of the lemma. Let $G$ denote the event that $\rvsigma^2_{\rvpi(k)}\geq d^2/4k^2$, and let $\rvtheta'_{\rvJ}=\frac{d}{2k}\rvtheta$. Then, conditional on $G$ we have that $\tilde\rvy_{i,\rvJ}\sim \cN\left(\rvtheta'_{\rvJ},\rvSigma_{\rvJ}\right)$ and thus, by the distribution of $\tilde\rvy_{i,\overline{\rvJ}}$, that $\tilde\rvY\sim\cN\left((\rvtheta'_{\rvJ},\rvtheta'_{\overline{\rvJ}}),\rvSigma\right)$. Additionally, by the distribution of $\rvtheta$ and $\rvtheta'_{\overline{\rvJ}}$, we have that $\rvtheta'=(\rvtheta'_{\rvJ},\rvtheta'_{\overline{\rvJ}})\sim\cN(0,\rho^2\id_d)$. Furthermore, conditional on $G$, for each $j\in[k]$ we have
$$
\abs{\cA_{\rvSigma}(\rvZ)_j-\rvtheta_j}=\frac{2k}{d}\abs{\cA(\tilde\rvY)_{\rvpi(j)} - \rvtheta'_{\rvpi(j)}}\equiv \frac{2k}{d}\abs{\cA(\rvY)_{\rvpi(j)} - \rvmu_{\rvpi(j)}},
$$
where the equality in distribution holds since conditional on $G$ we have $\rvY\equiv \tilde\rvY$ and $\rvmu\equiv\rvtheta'$. Thus, if $B_0$ denotes the event that $\abs{\set{j\in[k] : |\cA_{\rvSigma}(\rvZ)_j-\rvtheta_j|\leq \tau\cdot \frac{2k}{d}}}\geq k/2$ and $B_1$ denotes the event that $\abs{\set{j\in[k] : |\cA(\rvY)_{\rvpi(j)}-\rvmu_{\rvpi(j)}|\leq \tau}}\geq k/2$ then
$$
\Pr[B_1]\leq \Pr[\overline{G}] + \Pr[B_0].
$$
By our setting of $\rho$ and $\tau$ we have $2k\tau/d=2\sqrt{k}/Cn\leq 2k\rho/Cd$ (the standard deviation in \eqref{eq:k-dim-experiment} is $2k\rho/d$), and by our setting of $k$ and $n$, we have $k\geq C\paren{n + \frac{4k^2\tau^2n^2}{d^2}}$. Thus, we can apply \Cref{prop:coordinate-wise-error} to $\cA_{\rvSigma}$ with ambient dimension $k$ and coordinate-wise error threshold $\alpha\gets \tau\cdot \frac{2k}{d}$ to see that $\Pr[B_0]\leq \frac{1}{3}$. Additionally, since by \Cref{lem:top-k} we have that $\Pr[\overline{G}]\leq \frac{2}{k}$, we can conclude that $\Pr[B_1]\leq \frac{1}{3} + \frac{1}{12}$ for $k\geq 24$---that is, $\cA(\rvY)$ has error at most $\tau$ in at least $\frac{k}{2}$ of the coordinates in $\rvJ$ with probability at most $\frac 13 +\frac {1}{12}$.  

We now return to the analysis of mechanism $\cM$ (defined in \eqref{alg:diag:selection-reduction}). Applying the union bound over $B_1$ and the event controlled in \eqref{eq:boundedness-prob}, we see that with probability at least $\frac{1}{4}$ over the sampling of the instance $(\rvmu,\rvSigma)\sim\cP_{\text{diag}}$, the data $\rvX\sim\cN(0,\rvSigma)$, and the coins of the mechanism, $\cM$ outputs a vector $\rvs$ such that $\sum \rvs_i\leq \ell$, and $\rvs_j=1$ for at least $k/2$ of the coordinates $j\in\rvJ$.  Thus, $\cM$ is a $(k,\ell)$ selection mechanism with $\ell=\lceil\alpha^2/\tau^2\rceil$, which, by our setting of $\tau$, completes the proof.  
\end{proof}

\subsection{Private selection lower bound}
\label{sec:diag:selection-lb-proof}

To prove \Cref{thm:diag:selection-lb}, similarly to the diagonal case, we consider a test statistic that measures the correlation between the selection mechanism $\cM$ and the empirical covariance of the dataset. Fix a covariance matrix $\Sigma\in\R^{d\times d}=\mathrm{Diag}(\sigma^2_1,\dots,\sigma^2_d)$, a selection vector $s\in\{0,1\}^d$, and a test point $z\in\R^d$. Define the test
\begin{equation}
\label{eq:diag:test_def}
    \cT_{\Sigma,s}(z) := \sum_{j \in [d]}\cT_{\Sigma,s,j}(z), \quad \text{where}\quad \cT_{\Sigma, s, j}(z) := \paren{\frac{z_j^2}{\sigma_j^2}-1}\min\set{\sigma_j, L} s_j,
\end{equation}
and where $L=L(d,k)$ is a parameter which we will set later.  

We typically omit $\Sigma$ and $s$ from the notation for readability, using just $\cT$ and $\cT_j$.
Additionally, for convenience, we will use the shorthand $h(\sigma)$ for $\min\set{\sigma, L}$. Similarly to \cref{sec:non_diag}, we compare the distribution of the two statistics with covariance $\rvSigma\sim\cP_{\text{diag}}$, dataset $\rvX\sim\cN(0,\rvSigma)^{\otimes n}$, fresh sample $\rvz\sim\cN(0,\rvSigma)$, and selection vector $\rvs\sim\cM(\rvX)$:
\begin{itemize}
    \item The \textit{OUT sample case} in which we consider $\cT_{\rvSigma,\rvs}(\rvz)$ \item The \textit{IN sample case} in which we consider $\cT_{\rvSigma,\rvs}(\rvx_{\rvi})$ with $\rvi$ uniform in $[n]$. \end{itemize}

The following statements show that there exists a threshold $\tau\in(0,1)$ such that with high probability $\cT(\rvx_{\rvi})\geq d\tau/n$ and $\cT(\rvz)<d\tau/n$. We defer the proof of \Cref{lem:diag:out-case-bound} to \Cref{sec:diag:out-case-analysis}, and \Cref{lem:diag:in-case-bound} to \Cref{sec:diag:in-case-analysis}, and complete the proof of the selection lower bound (\Cref{thm:diag:selection-lb}) below. 

\begin{lemma}[OUT sample test is small]
    \label{lem:diag:out-case-bound} 
    Fix $\tau\in(0,1)$. There exist $C>0$ such that for all $d,k,n,\ell,\in\N$ with $k \ge C n^2\log n$, if $L=\frac{d}{4k}$ then every $(k,\ell)$-selection mechanism $\cM$ satisfies 
    \[
    \Pr_{\rvSigma,\rvX,\rvz,\cM}\brackets{\cT(\rvz)\geq \frac{d\tau}{n}} = O\paren{\frac{1}{n^2}}.
    \]
\end{lemma}

\begin{lemma}[IN sample test is large]
    \label{lem:diag:in-case-bound} 
    Let $C>0$ be sufficiently large. There exists $\tau\in(0,1)$ such that for all $d,k,n,\ell\in\N$ with $C\max\set{n,\log d}\leq k\leq d/C$ and $\ell\leq d/C\log d$, if $L=\frac{d}{4k}$ then every $(k,\ell)$-selection mechanism $\cM$ satisfies   
    \[
    \Pr_{\rvSigma,\rvX,\rvi,\cM}\brackets{\cT(\rvx_{\rvi})\geq \frac{d\tau}{n}} = \Omega\paren{\frac1n}.
    \]
\end{lemma}

\begin{proof}[Proof of \Cref{thm:diag:selection-lb}]
The proof follows by contradiction. Suppose $\cM$ is an $(\eps,\delta)$-DP $(k,\ell)$-selection mechanism with $\eps\leq 1$, $\delta\leq c/n$, $k\geq C\max\set{n^2\log n, \log d}$, and $\ell\leq d/C\log(d)$.

For every index $i\in [n]$ let $\rvX^{\oplus i}$ denote $\rvX$ with $\rvx_i$ replaced by $\rvz$, and let $\cT^{\oplus i}$ denote $\cT$ executed on $\cM(\rvX^{\oplus i})$. Then, since $\cM$ is $(\eps,\delta)$-DP we have 
\[
\cT(\rvz)\approx_{\eps,\delta} \cT^{\oplus i}(\rvz)\equiv \cT(\rvx_i)
\]
where $\approx_{\eps,\delta}$ is defined in \Cref{def:indistinguishability}, and the equality in distribution follows from the fact that $\rvx_i$ and $\rvz$ are both sampled from $\cN(0,\Sigma)$ independently of the other datapoints. Hence, if $\rvi\sim[n]$, and $\tau$ is the constant in \Cref{lem:diag:in-case-bound}, then
\[
\Omega\paren{\frac{1}{n}}\leq \Pr\brackets{\cT(\rvx_{\rvi})\geq \frac{d\tau}{n}}\leq e^{\eps}\Pr\brackets{\cT(\rvz)\geq \frac{d\tau}{n}} + \delta\leq O\paren{\frac{1}{n^2}} + \frac{c}{n}.
\]
which, for sufficiently small $c$ yields a contradiction. 
\end{proof}

\subsubsection{Analysis of the test in the OUT case}
\label{sec:diag:out-case-analysis}

In this section, we prove \Cref{lem:diag:out-case-bound}, which states that $\cT(\rvz)\geq d\tau/n$ with probability at most $O(1/n^2)$.

\begin{proof}[Proof of \Cref{lem:diag:out-case-bound}] First we bound the expectation. Observe that for every fixed $\Sigma$, $X$, and $s=\cM(X)$ 
    $$
    \Ex_{\rvz_j}[\cT_j(\rvz)] = \Ex_{\rvz_j}\left[\frac{\rvz_j^2}{\sigma_j^2}-1\right]h(\sigma_j)s_j = 0
    $$
    and thus, by linearity of expectation $\Ex[\cT(\rvz)]=0$. 

    Next, we prove  a high probability bound on $\cT(\rvz)$. Since for fixed $\sigma_j$ and $\rvz_j\sim\cN(0,\sigma_j^2)$, the random variable $\rvz_j^2/\sigma^2_j$ is $\chi^2_{1}$, we can apply a scalar Bernstein inequality (e.g., Corollary 2.9.2 \cite{Vershynin26}) to see that 
    $$
    \Pr\left[ \abs{\sum_{j\in[d]}\cT_j(\rvz)} \geq  \sqrt{\sum_{j\in [d]}h(\sigma_j)^2\log(1/\beta)s_j} + L\log(1/\beta) \right]\leq \beta.
    $$
    Next, we prove a high probability bound on $\sum h(\rvsigma_j)^2\rvs_j$.  

    The proof goes through a Chernoff bound for bounded nonnegative random variables (\Cref{lem:bounded-rv-chernoff}) applied to $h(\rvsigma_j)^2\in [0,L^2]$. We first bound the expectation:
    $$
    \Ex[h(\rvsigma_j)^2] \leq \int_{0}^{L^2}\Pr[\rvsigma_j^2\geq t]dt\leq 1+\int_{1}^{L^2}t^{-1/2}dt\leq 2L 
    $$
    Applying \Cref{lem:bounded-rv-chernoff},
    there exists a constant $C_1$ such that
    $$
    \sqrt{\sum_{j\in [d]}h(\rvsigma_j)^2\log(1/\beta)}\leq \sqrt{C_1Ld\log(1/\beta)} + L\log(1/\beta)
    $$
    with probability at least $1-\beta$, which, since $\rvs_j\in\{0,1\}$ implies the same bound on the sum involving $\rvs_j$. Setting $\beta=O(1/n^2)$ and choosing $C=C(\tau, C_1)$, we see that if $L=d/4k\leq d/Cn^2\log n$ we obtain the desired bound
    $$
    \Pr\left[\sum_{j\in[d]}\cT_j(\rvz)\geq \frac{d\tau}{n}\right]\leq O\left(\frac{1}{n^2}\right).
    $$  
\end{proof}

\subsubsection{Analysis of the test in the IN case}
\label{sec:diag:in-case-analysis}

This section is devoted to the proof of \Cref{lem:diag:in-case-bound}, which states that $\cT(\rvx_{\rvi})\geq d\tau/n$ with probability at least $\Omega(1/n)$. Throughout this section we let $\Ex[\cT_j(\rvx_{\rvi})\mid \rvX,\rvs]$ denote the random variable given by sampling a realization $X$ and $s$ of $\rvX$ and $\rvs$, and returning $\Ex[\cT_j(x_{\rvi})\mid \rvX=X,\rvs=s]$. Additionally, for random variables $\rvw$ and $\rvk$, let $\rvw\cond{\rvk}$ denote the random variable given by sampling a realization $k$ of $\rvk$ and then sampling a realization $w$ from the distribution of $\rvw$ conditioned on the event that $\rvk=k$. Before formally stating the claims and proofs, we provide some intuition for the behavior of $\Ex[\cT_j(\rvx_{\rvi})\mid \rvX,\rvs]$ for fixed $j\in [d]$. It will be convenient to use the following conjugate prior property of the scaled-inverse chi-squared distribution:

\begin{fact}[Conjugate prior property]
\label{fact:invchi_properties}
    Let $\rvsigma^2 \sim \invchi(\nu,\tau^2)$, and $\rvX = (\rvx_1,\ldots,\rvx_n) \sim \cN(0,\rvsigma^2)^{\otimes n}$, then it holds that: 
    \[
    \rvsigma^2\cond{\rvX} \sim \invchi\paren{\nu + n, \frac{\nu \tau^2 + \sum_{i \in [n]} \rvx_i^2}{\nu + n}}.
    \]
\end{fact}
Let $\rvq_{j}=\frac{1}{n}\sum_{i\in[n]}\rvx_{i,j}^2$ denote the empirical variance of $\rvX$ in coordinate $j$. Since $\rvsigma^2_j\sim\invchi(1,1)$, by \Cref{fact:invchi_properties},  conditional on $\rvX$ we have $\rvsigma^2_j\sim\invchi\paren{n+1, \frac{1+n\rvq_j}{n+1}}$. Assuming $n$ is large, we have that $\rvsigma_j^2\approx \frac{1+n\rvq_j}{1+n}$ with high probability. Thus, we expect that
\begin{align*}
\Ex[\cT_j(\rvx_{\rvi})\mid \rvX,\rvs]
&= \Ex\left[\paren{\frac{\Ex_{\rvi}[\rvx_{\rvi,j}^2]}{\rvsigma^2_j}-1}h(\rvsigma_j)\rvs_j~\middle|~\rvX,\rvs\right]\\
&\approx \frac{\rvq_j-1}{1+n\rvq_j}\min\paren{\sqrt{\frac{1+n\rvq_j}{1+n}},L}\rvs_j, 
\end{align*}
where the first equality follows since $\rvi$ is independent of $\rvX$ and $\rvSigma$, and the second equality follows since $\Ex[\rvx_{\rvi,j}^2\mid \rvX]=\rvq_j$, and by substituting the approximation $\frac{1+n\rvq_j}{1+n}$ for $\rvsigma^2_j$. Inspecting the expression on the right hand side, we see that (ignoring constant factors) there are two salient regimes: First, when $\rvq_j \gg 1$, the expectation is at least $\min\paren{\sqrt{\rvq_j},L}/n$, and second, when $\rvq_j \ll 1$, algebraic manipulation shows that the expectation is always at least $-\sqrt{\rvq_j}/n$. We therefore expect the expectation to have the following behavior:
$$
\Ex[\cT_j(\rvx_{\rvi})\mid \rvX,\rvs] \approx  \frac{\rvs_j}{n}\cdot 
\begin{cases}
    \min\paren{\sqrt{\rvq_j},L} & \rvq_j > 1\\
    -\sqrt{\rvq_j} & \rvq_j\leq 1
\end{cases}
$$
Subject to the above heuristic, we see that for each coordinate $j$ with $\rvq_j\leq 1$, the expectation is always at least $-1/n$. Since $\cM$ can select at most $\ell$ coordinates, these coordinates can contribute a total of $\frac{-\ell}{n}$ to the sum. On the other hand, if the coordinates with the $k$ largest $\rvq_j$ are all at least $L$, then, because $\cM$ is required to select at least a constant fraction of them, they must contribute at least $\frac{kL}{n}$ to the sum. It follows that the expectation is large whenever $kL\gg \ell\approx d/\log d$, or equivalently when $L\approx d/k$. Using the distribution of $\rvsigma_j$ and $\rvX$, one can show that the $k$ largest $\rvq_j$ are indeed typically at least $d/k$, which completes proof. 

The first step in our proof is \Cref{claim:diag:expectation-formula}, which provides a more tractable form for analyzing the expectation of the test conditional on $\rvX$ and $\rvs$. While the heuristic analysis above provides intuition by reasoning about the empirical variance $\rvq_j$, \Cref{claim:diag:expectation-formula} provides an equivalent representation of the expectation over $\rvsigma_j$ that exhibits the same qualitative behavior as the heuristic. 

\begin{claim}[Simplifying the expectation]
    \label{claim:diag:expectation-formula}
    For all $j\in[d]$, the conditional expectation of $\cT_j$ satisfies
    $$
    \Ex\left[\cT_j(\rvx_{\rvi})\mid \rvX,\rvs\right]=\frac{1}{n}\cdot \Ex\left[\paren{L\Ind[\rvsigma_j>L] - \frac{h(\rvsigma_j)}{\rvsigma_j^2}}\rvs_j~\middle|~ \rvX,\rvs\right].
    $$
    In particular,
    $$
    \Ex[\cT(\rvx_{\rvi})\mid \rvX,\rvs] = \frac{1}{n}\cdot \Ex\left[\sum_{j\in[d]} \paren{L\Ind[\rvsigma_j>L] - \frac{h(\rvsigma_j)}{\rvsigma_j^2}}\rvs_j~\middle|~\rvX,\rvs\right].
    $$
\end{claim}
\begin{proof}
    Since $\rvSigma$ is independent of $\rvs$ conditional on $\rvX$, we first analyze the expectation over $\rvSigma$ conditional on $\rvX$, and then include the conditioning on $\rvs$ at the end of the proof. Let $\rvq_j=\frac{1}{n}\sum_{i\in[n]} \rvx_{i,j}^2$ be the empirical variance in coordinate $j$. The proof follows from integration by parts. In what follows, we consider a fixed coordinate $j\in[d]$, and omit the subscript $j$ from the calculations to reduce notational clutter. We begin by showing the following equation: 
    \begin{align}
        \label{eq:diag:integration-by-parts}
        \Ex[\rvsigma h'(\rvsigma)\mid \rvX] = \Ex\left[h(\rvsigma)\paren{(n+1) - \frac{1+n\rvq}{\rvsigma^2}}~\middle|~ \rvX\right].
    \end{align}
    Recall by \Cref{fact:invchi_properties}, the distribution of $\rvsigma^2|_{\rvX}$ has density $p_{\rvsigma^2}(t)\propto t^{-(\nu/2 + 1)}\exp\paren{\frac{-\nu\tau^2}{2t}}$, where $\nu=n+1$ and $\tau^2=\frac{1+n\rvq}{1+n}$. Since the quantity in the expectation in question is $\rvsigma$ (as opposed to $\rvsigma^2$), it will be convenient to work with the density of $\rvsigma|_{\rvX}$ instead. Let $f_{\rvsigma}(s)$ be the density of $\rvsigma|_{\rvX}$. Then, by a change of measure, we have that $f_{\rvsigma}(s)=2s p_{\rvsigma^2}(s^2)\propto s^{-(\nu+1)}\exp\paren{\frac{-\nu\tau^2}{2s^2}}$. Now, we apply integration by parts to the expectation:
    \begin{align*}
        \Ex[\rvsigma h'(\rvsigma)\mid \rvX]
        & =\int_{s=0}^{\infty} s h'(s) f_{\rvsigma}(s)ds \\
        & = \left[ sf_{\rvsigma}(s)h(s)\right]_{0}^{\infty} - \int_{s=0}^{\infty} h(s) (sf_{\rvsigma}(s))'ds.
    \end{align*}
    Since $h(s)\leq L$ and $\lim_{s\to \infty} sf_{\rvsigma}(s) = \lim_{s\to 0} sf_{\rvsigma}(s) = 0$, the boundary term is $0$. To handle the remaining term, we use the formula for $f_{\rvsigma}(s)$ to explicitly compute the derivative: 
    $$
    (sf_{\rvsigma}(s))' = f_{\rvsigma}(s)\paren{\frac{1+n\rvq}{s^2}-(n+1)}. 
    $$
    Substituting in the integral $\int h(s)(sf_{\rvsigma}(s))'ds$ yields \eqref{eq:diag:integration-by-parts}. By algebra, we see that 
    $$
    (n+1) - \frac{1+n\rvq}{\rvsigma^2} = \paren{1-\frac{1}{\rvsigma^2}} - n\paren{\frac{\rvq}{\rvsigma^2}-1},
    $$
    and thus, that
    $$
    n\Ex\left[\paren{\frac{\rvq}{\rvsigma^2}-1}h(\rvsigma)~\middle|~\rvX \right]=\Ex\left[h(\rvsigma) - \rvsigma h'(\rvsigma) - \frac{h(\rvsigma)}{\rvsigma^2} ~\middle|~\rvX\right].
    $$
    The statement follows since $h(\rvsigma)-\rvsigma h'(\rvsigma) = L\Ind[\rvsigma > L]$, and since 
    $$
    \Ex\left[\paren{\frac{\rvq_j}{\rvsigma_j^2} - 1}h(\rvsigma_j)~\middle|~\rvX\right]\rvs_j = \Ex\left[\Ex_{\rvi\sim[n]}\left[\paren{\frac{\rvx_{\rvi,j}^2}{\rvsigma_j^2} - 1}h(\rvsigma_j)\right]~\middle|~\rvX\right]\rvs_j=\Ex[\cT_j(\rvx_{\rvi})\mid\rvX,\rvs],
    $$ 
    which holds because $\rvi\sim [n]$ uniformly and independently of $\rvX$, $\rvs$, and $\rvSigma$.  
\end{proof}

In order to show that the conditional expectation is large, we must show that the positive part of the expression in \Cref{claim:diag:expectation-formula} is much larger than the negative part. In line with our heuristic derivation, \Cref{claim:diag:positive-terms-lb} shows that when $L=d/4k$, the positive part is $\Omega(d)$ with constant probability. 

\begin{claim}[Lower bounding the positive part]
    \label{claim:diag:positive-terms-lb}
    There exists $c>0$ such that if $L=d/4k$ and $k\in \left[\max(6\log d,24), cd\right]$ then 
    $$
    \Ex\left[\sum_{j\in[d]} L\Ind[\rvsigma_j>L]\rvs_j~\middle|~ \rvX,\rvs\right]\geq cd 
    $$
    with probability at least $c$.
\end{claim}

Before diving in to the proof, we give a brief overview of the main ideas. Let $\rvg = \sum L\Ind[\rvsigma_j>L]\rvs_j$ and $\mu_{\rvX,\rvs} = \Ex[\rvg\mid \rvX,\rvs]$. Then \Cref{claim:diag:positive-terms-lb} states that $\mu_{\rvX,\rvs}\geq cd$ with constant probability. The essence of the proof is the following three observations: First, $\rvg$ is bounded above by $Ld$. Second, the expectation $\Ex[\mu_{\rvX,\rvs}]=\Ex[\rvg]$ is in the interval $[cd,2d]$. And third, since $\rvg$ is bounded above by a sum of i.i.d. Bernoulli random variables, $\Pr[\rvg\geq 4d]\leq e^{-d/3L}$. Taken together, these observations imply that $\mu_{\rvX,\rvs}$ cannot place too much mass above $4d$, and, since $\Ex[\mu_{\rvX,\rvs}]\in[cd,2d]$, it must place sufficient mass above $cd/2$.

\begin{proof}
    As in the proof sketch above, let $\rvg = \sum L\Ind[\rvsigma_j>L]\rvs_j$ and $\mu_{\rvX,\rvs} = \Ex[\rvg\mid \rvX,\rvs]$. We first prove the bound on the expectation. The upper bound follows since $\rvs_j=\cM(\rvX)_j\leq 1$, and since $\Pr[\rvsigma_{j}\geq L]\leq 2/L$ (because $\rvsigma^2\sim\invchi(1,1)$). To see why the lower bound holds, observe that by \Cref{lem:top-k} we have that $\rvsigma_{(k)}\geq d/4k$ with probability at least $1-2/k$. Let $\bm{\pi}$ be the permutation ordering $(\rvsigma^2_j)_{j\in[d]}$. Since $\rvs_{\bm{\pi}(j)}=1$ for at least $k/4$ of the $j\in[k]$ with probability at least $1/3$, the union bound yields
    \[
    \sum_{j\in[k]} L\Ind[\rvsigma_{\bm{\pi}(j)}>L]\rvs_{\bm{\pi}(j)}\geq \sum_{j\in{[k]}} \frac{d}{4k}\rvs_{\bm{\pi}(j)}\geq \frac{d}{16}
    \]
    with probability at least $1/4$ for $k\ge 24$, and thus, $\Ex[\mu_{\rvX,\rvs}]=\Ex[\rvg]\in [cd, 2d]$ for some constant $c>0$. 

    Next, we prove a high probability bound on $\rvg$. Observe that $\rvg\leq \sum L\Ind[\rvsigma_j>L]$, which is a sum of i.i.d. Bernoulli random variables with mean at most $2d$. Let $B_1$ denote the event that $\rvg\geq 4d$, and let $B_2$ denote the event that $\Ex[\Ind[B_1]\mid \rvX,\rvs]\geq e^{-d/6L}$. By a Chernoff bound, $\Pr[B_1]\leq e^{-d/3L}$, and thus, by Markov's inequality, $\Ex[\Ind[B_2]]\leq e^{-d/6L}$. In other words, $\rvg$ is rarely bigger than $4d$, and with high probability over $\rvX$ and $\rvs$, random variable $\rvg|_{\rvX,\rvs}$ is rarely bigger than $4d$. 

    We now argue that $\mu_{\rvX,\rvs}$ must be at least $\Omega(d)$ with constant probability over $\rvX$ and $\rvs$.

    \begin{align*}
        cd 
        \leq \Ex\brackets{\mu_{\rvX,\rvs}}
        &\leq \Ex[\mu_{\rvX,\rvs}\Ind[\overline{B_2}]] + Lde^{-d/6L}\\
        &\leq \Ex\left[\Ex[\rvg\Ind[\overline B_1]\mid\rvX,\rvs]\right] + \underbrace{Ld\Ex\left[\Ex[\Ind[B_1]\mid\rvX,\rvs]\Ind[\overline{B_2}]\right]}_{\leq Lde^{-d/6L}} + Lde^{-d/6L}\\
        & \leq \Ex[\mu_{\rvX,\rvs}\Ind[\mu_{\rvX,\rvs}<c'd]] + 4d\Ex[\Ind[\mu_{\rvX,\rvs}>c'd]] + 2Lde^{-k/6}~~~\text{for any $c'>0$}\\
        &\leq 2c'd + 4d\Pr[\mu_{\rvX,\rvs}>c'd],
    \end{align*}
    where the final inequality holds since $L=d/4k$ and $k\geq 6\log(d)$.  Setting $c'=c/4$ and rearranging terms, we see that $\mu_{\rvX,\rvs}\geq cd/4$ with probability at least $c/8$.
\end{proof}

To complete our analysis of the conditional expectation, we prove \Cref{claim:diag:negative-terms-ub}, which states that the negative terms of the conditional expectation cannot decrease the overall sum by much more than $\sqrt{\ell d}$. We note that this analysis diverges slightly from the heuristic bound of $\ell$ derived at the start of the section; however, the deviation is minor---for $\ell\approx d/\log d$ we have that $\sqrt{\ell d}\approx \ell\sqrt{\log d}$. 

\begin{claim}[Upper bounding the negative part]
    \label{claim:diag:negative-terms-ub}
    For all $C>0$
    $$
    \Ex\left[\sum_{j\in[d]} \frac{h(\rvsigma_j)}{\rvsigma_j^2}\rvs_j~\middle|~ \rvX,\rvs\right]\leq C\sqrt{\ell d} 
    $$
    with probability at least $1-1/C$.
\end{claim}
\begin{proof}
    Since $h(\rvsigma_j)\leq \rvsigma_j$, and $\Ex[\rvsigma^{-2}_j]< \infty$, we can bound the expectation of the negative terms over $\rvX$ and $\rvs$ using Cauchy-Schwarz and Jensen's inequality
    $$
    \Ex\left[\sum \frac{h(\rvsigma_j)}{\rvsigma_{j}^2}\rvs_j\right]\leq \Ex\brackets{\sqrt{\ell\sum \rvsigma^{-2}_j}}\leq \sqrt{\ell\sum \Ex[\rvsigma_j^{-2}]}= \sqrt{\ell d},
    $$
    where the final inequality follows because $\rvsigma_j^{-2}$ follows a $\chi^2_{1}$ distribution. By Markov's inequality over $\rvX$ and $\rvs$, the negative terms are at most $C\sqrt{\ell d}$ with probability at least $1-1/C$.
\end{proof}

\Cref{claim:diag:expectation-formula,claim:diag:positive-terms-lb,claim:diag:negative-terms-ub} show that the conditional expectation is $\Omega(d/n)$ with constant probability. However, in order to show that the test itself is large with constant probability (over the randomness of both $\rvSigma$ and $\rvX$), we require a bound on the conditional variance. \Cref{claim:diag:variance-bound} states that the variance of the test conditional on $\rvX$ is typically less than $Ld/n$. 

\begin{claim}[Bounding the variance]
    \label{claim:diag:variance-bound}
    For all $C>0$
    $$
    \Var\left[\Ex\left[\cT(\rvx_{\rvi})~\middle|~ \rvSigma\right]~\middle|~ \rvX,\rvs\right]\leq \frac{4CLd}{n}
    $$
    with probability at least $1-1/C$.
\end{claim}
\begin{proof}
     Since the $\rvsigma^2_j$'s are sampled independently for each $j\in[d]$, we have that conditioned on $\rvX$ and $\rvs$, the terms $\cT_j(\rvx_{\rvi})$ are independent. Thus, 
    \begin{align*}
    \Var_{\rvSigma}\left[\Ex_{\rvi}\brackets{\cT(\rvx_{\rvi})\mid \rvSigma}~\middle|~ \rvX,\rvs\right] 
    &= \sum_{j\in [d]}\Var\left[ \paren{\frac{\rvq_j}{\rvsigma_j^2}-1}h(\rvsigma_j)\rvs_j ~\middle|~ \rvX,\rvs \right]\\
    & \leq \sum_{j\in [d]}\Ex_{\rvSigma}\left[ \paren{\frac{\rvq_j}{\rvsigma_j^2}-1}^2h(\rvsigma_j)^2 ~\middle|~ \rvX \right].
    \end{align*}
    Taking the expectation over $\rvX$ and $\rvs$ and swapping the order of expectation we see that,
    \begin{align*}
    \Ex_{\rvX,\rvs}\left[ \Var_{\rvSigma}\left[\Ex_{\rvi}\left[\cT(\rvx_{\rvi})\mid \rvSigma\right]~\middle|~ \rvX,\rvs\right]\right] 
    & \leq \sum_{j\in [d]}\Ex_{\rvSigma}\Ex_{\rvX}\left[ \paren{\frac{\rvq_j}{\rvsigma_j^2}-1}^2h(\rvsigma_j)^2 ~\middle|~\rvSigma\right]\\
    & \leq \sum_{j\in [d]}\Ex_{\rvSigma}\left[ \frac{2}{n}\cdot h(\rvsigma_j)^2 \right]\\
    &\leq \frac{4dL}{n}.
    \end{align*}
    The last inequality follows from the fact that $\Ex[h(\rvsigma_j)^2]\leq 2L$ (see the proof of \Cref{lem:diag:out-case-bound}). Applying Markov's inequality completes the proof of the claim. 
\end{proof}

We now complete the proof of our main lemma, which says that $\Pr[\cT(\rvx_{\rvi})\geq d\tau/n]=\Omega(1/n)$. 

\begin{proof}[Proof of \Cref{lem:diag:in-case-bound}] Let $c>0$ be the constant from \Cref{claim:diag:positive-terms-lb}, and define $G$ as the event that 
$$
\Ex[\cT(\rvx_{\rvi})\mid\rvX,\rvs]\geq \frac{cd}{2n} ~~\text{and}~ \Var[\Ex[\cT(\rvx_{\rvi})\mid \rvSigma]\mid \rvX,\rvs]\leq \frac{16Ld}{cn}.
$$
Let $\rvmu_{\rvSigma}=\Ex[\cT(\rvx_{\rvi})\mid \rvSigma,\rvX,\rvs]$, and let $E$ be the event that 
$$
\rvmu_{\rvSigma}\geq \frac{1}{2}\Ex[\rvmu_{\rvSigma}\mid \rvX,\rvs]=\frac{1}{2}\Ex_{\rvSigma,\rvi}\brackets{\cT(\rvx_{\rvi})\mid \rvX,\rvs}.
$$
Since conditional on $G$ we have $\Ex[\cT(\rvx_{\rvi})\mid \rvX,\rvs]>0$, we can apply Paley-Zygmund conditional on $G$ to obtain that, 
\begin{align*}
\Ex_{\rvSigma}[\Ind_{E}\mid \rvX,\rvs]\cond{G}
&\geq \frac{1}{4}\frac{\Ex_{\rvSigma}[\rvmu_{\rvSigma}\mid \rvX,\rvs]\cond{G}^2}{\Ex_{\rvSigma}[\rvmu_{\rvSigma}^2\mid \rvX,\rvs]\cond{G}}\\
&=\frac{1}{4}\frac{\Ex_{\rvSigma}[\rvmu_{\rvSigma}\mid \rvX,\rvs]\cond{G}^2}{\Var[\rvmu_{\rvSigma}\mid \rvX,\rvs]\cond{G} + \Ex_{\rvSigma}[\rvmu_{\rvSigma}\mid \rvX,\rvs]\cond{G}^2}\\
&\geq\frac{1}{6}
\end{align*}
where the final inequality follows since by definition of $G$ and $\rvmu_{\rvSigma}$, and our setting of $k$,
$$
\frac{\Var[\rvmu_{\rvSigma}\mid \rvX,\rvs]\cond{G}}{\Ex_{\rvSigma}[\rvmu_{\rvSigma}\mid \rvX,\rvs]\cond{G}^2} 
= \frac{\Var[\Ex[\cT(\rvx_{\rvi})\mid \rvSigma]\mid \rvX,\rvs]\cond{G}}{\Ex_{\rvSigma,\rvi}[\cT(\rvx_{\rvi})\mid \rvX,\rvs]\cond{G}^2}\leq \frac{16n}{c^3 k}
\leq \frac{1}{2}.
$$
Taking the expectation over $\rvs$ and $\rvX$, we see that $\Ex_{\rvX,\rvs,\rvSigma}[\Ind_{E}\mid G]\geq 1/6$.

Next we argue that $\Pr[G]\geq c/2$. By \Cref{claim:diag:expectation-formula}, the random variable $\Ex[\cT(\rvx_{\rvi})\mid  \rvX,\rvs]$ can be decomposed into a positive part and a negative part. By \Cref{claim:diag:positive-terms-lb}, the positive part is at least $cd$ with probability at least $c$, and by \Cref{claim:diag:negative-terms-ub}, the negative part is at most $4c^{-1}\sqrt{\ell d}$ with probability at least $1-c/4$, which, since $\ell\leq d/C\log d$ and $C>0$ is large enough, is at most $cd/2$. If both of these events occur then the expectation is at least $cd/2n$. Additionally, by \Cref{claim:diag:variance-bound}, the variance is at most $16c^{-1}Ld/n$ with probability at least $1-c/4$. By the union bound $G$ occurs with probability at least $c/2$. 

To complete the proof, fix $\tau\in(0,c/4)$ and let $A_1$ be the event that $\cT(\rvx_{\rvi})\geq d\tau/n$. There exists $\rvw=\rvw(\rvX,\rvs,\rvSigma)\in[n]$ such that $\cT(\rvx_{\rvw})\geq \Ex_{\rvi}[\cT(\rvx_{\rvi})\mid \rvX,\rvs,\rvSigma]$. Let $A_2$ be the event that $\rvi=\rvw$. Then conditional on $G$ we have $\Ind_{A_1}\Ind_{A_2}\geq \Ind_{E}\Ind_{A_2}$. Hence, 
\begin{align*}
\Pr_{\rvX,\rvs,\rvSigma,\rvi}\brackets{\cT(\rvx_{\rvi})\geq \frac{d\tau}{n}} 
& \geq \Ex_{\rvX,\rvs,\rvSigma,\rvi}[\Ind_{A_1}\Ind_{A_2}]\\
& \geq \frac{c}{2}\Ex_{\rvX,\rvs,\rvSigma}[\Ind_{E}\Ex_{\rvi}[\Ind_{A_2}\mid \rvX,\rvs,\rvSigma]\mid G]\\
& \geq\frac{c}{2n}\Ex_{\rvX,\rvs,\rvSigma}[\Ind_{E}\mid G]=\frac{c}{12n},
\end{align*}
where the second inequality follows since $\Pr[G]\geq c/2$, and since $\Ind_{E}$ is independent of $\rvi$, and the third inequality follows since $\Pr[\rvi=\rvw]=1/n$.  
\end{proof}

\section{Lower bound for mean estimation with Mahalanobis error}
\label{sec:rel}

In this section we argue that private mean estimation in $d$ dimensions with error relative to the covariance $\Sigma$ is impossible with less than $d$ samples. Let $\Sigma^{\dagger}$ denote the pseudo-inverse of a positive semidefinite matrix $\Sigma$. For a vector $y\in\R^d$, the Mahalanobis norm of $y$ relative to $\Sigma$ is defined by
$$
d_{\Sigma}(y) = 
\begin{cases}
    \sqrt{y^\top\Sigma^{\dagger}y} & y\in\Span(\Sigma)\\
    \infty & \text{otherwise}
\end{cases}
$$
Our lower bound uses the following prior $\cP_{\text{rel}}$ over instances $(\rvmu,\rvSigma)$: 
\begin{align}
    \label{eq:rel:prior}
    \paren{\rvmu,\rvSigma}\sim\cP_{\text{rel}}(d) \iff 
    \begin{cases}
        \rvmu\sim\cN(0,\id_d)  \\\rvSigma = \id_d-\rvv\rvv^\top \ \text{where $\rvv\sim\set{v : \|v\|_2=1}$ uniformly.}
    \end{cases}
\end{align}
We note that for instances $\rvSigma$ in the support of $\cP_{\text{rel}}$ and any vector $y\in\R^d$ we either have $d_{\rvSigma}(y)=\|y\|_2$ if $y\in\Span(\rvSigma)$, or $d_{\rvSigma}(y)=\infty$ otherwise.

\begin{theorem}
\label{thm:rel:estimation-lb}
There exists a constant $C>0$ such that for all $d,n\in\N$ and $\eps,\delta\geq 0$, if $d\geq n+2$ then every $(\eps,\delta)$-DP mechanism $\cM$ satisfies
$$
\Pr_{\substack{(\rvmu,\rvSigma)\sim\cP_{\text{rel}} \\ \rvX\sim\cN(\rvmu,\rvSigma)^{\otimes n}}}\brackets{d_{\rvSigma}\paren{\cM(\rvX)-\rvmu} < \infty}\leq n\delta.
$$
\end{theorem}

The intuition behind the proof, is that conditioned on $\rvX-\rvmu$, the span of $\rvSigma$ is composed of $\Span(\rvX-\rvmu)$ and a uniformly random $d-n-1$ dimensional subspace of $\Span(\rvX-\rvmu)^{\perp}$. If the algorithm's output is not in $\Span(\rvX-\rvmu)$, then with probability $1$ over the posterior sampling of $\rvv$, its Mahalanobis error is infinite. Thus, the algorithm's output must be within $\Span(\rvX-\rvmu)$, from which we derive a privacy violation. 

\begin{proof}
We show a contradiction to $(\eps,\delta)$-DP by constructing an attack that distinguishes between $\rvz\sim\cN(\rvmu,\rvSigma)$ and $\rvx_{\rvi}$, where $\rvi\sim[n]$ uniformly. The test uses the output of the mechanism $\rvy$, as well as the other $n-1$ datapoints $\rvX_{-\rvi}$. For all $z,\mu\in\R^d$ and datasets $X\in(\R^d)^{n-1}$ define 
$$
S_{X,\mu}(z)=\Span\set{x_1-\mu,\dots,x_{n-1}-\mu,z-\mu}\setminus\Span\set{x_1-\mu,\dots,x_{n-1}-\mu}.
$$
And define the test $T_{X,\mu,y}(z)=\Ind[y-\mu\in S_{X,\mu}(z)]$. In the remainder of the proof, we will omit the subscripts $X,\mu$ and $y$, since they will be clear from context.\\

Let $(\rvmu,\rvSigma)\sim\cP_{\text{rel}}$, $\rvX\sim\cN(\mu,\rvSigma)^{\otimes n}$, $\rvy\sim\cM(\rvS)$, $\rvz\sim\cN(\rvmu,\rvSigma)$, and $\rvi\sim[n]$ uniformly and independently.  Let $G$ be the event that $d_{\rvSigma}(\rvy-\rvmu)<\infty$ and let $E$ be the event that $\rvy-\rvmu\in\Span(\rvX-\rvmu)$. The key idea behind our proof is that $\Pr[G]=\Pr[E]$, that is, the probability that the Mahalanobis error is finite is exactly the probability that $\rvy-\rvmu$ is in the span of $\rvX-\rvmu$. To see why this is, observe that 
\begin{align*}
\Ex[\Ind_G]
&=\Ex[\Ind_G\Ind_E] + \Ex[\Ind_G\Ind_{\overline{E}}] \\
&= \Ex[\Ind_E] + \Ex_{\rvX,\rvy,\rvmu}[\Ex_{\rvv}[\Ind_G\mid \rvX,\rvy,\rvmu]\Ind_{\overline{E}}]\\
&= \Ex[\Ind_E]
\end{align*}
where the second equality follows since $\Ind_E=1$ implies $\Ind_G=1$. The final equality follows since conditional on $\rvy,\rvX$, and $\rvmu$ such that $\Ind_{\overline E}=1$, over the random sampling of the direction $\rvv$ (see \eqref{eq:rel:prior}), we have $\langle \rvv, \rvy-\rvmu\rangle\neq 0$ with probability $1$, and thus $d_{\rvSigma}(\rvy-\rvmu)=\infty$, implying $\Ex_{\rvv}[\Ind_G\mid \rvX,\rvy,\rvmu]=0$. 

\paragraph{IN case.} Since $n<d$ the data $\set{\rvx_i-\rvmu}_{i\in[n]}$ are linearly independent with probability $1$, and thus $\bigcap_{i\in[n]} \Span(\rvX_{-i}-\rvmu)=\set{0}$. It follows that there exists some random variable $\rvw\in[n]$, such that when $\rvy\neq \rvmu$ and conditional on $E$, we have $\rvy-\rvmu\not\in\Span\paren{\rvX_{-\rvw}-\rvmu}$, or equivalently $T_{\rvX_{-\rvw},\rvmu}(\rvx_{\rvw})=1$, with probability $1$. Since $\rvi=\rvw$ with probability $1/n$ independently of $\rvy,\rvX,\rvu$ and $\rvSigma$, we obtain, 
\begin{align*}
\Pr[T_{\rvX_{-\rvi},\rvmu}(\rvx_{\rvi})=1]
&\geq \frac{1}{n}\Pr[T_{\rvX_{-\rvw},\rvmu}(\rvx_{\rvw})=1\mid E\cap \set{\rvy\neq \rvmu}]\Pr[E\cap \set{\rvy\neq\rvmu}]\\
&\geq \frac{\Pr[E]-\Pr[\rvy=\rvmu]}{n}=\frac{\Pr[G]}{n},
\end{align*}
where the final equality holds since $\Pr[\rvy=\rvmu]=0$ and $\Pr[E]=\Pr[G]$.

\paragraph{OUT case.} Since $n\leq d-2$, for fixed $\rvy,\rvmu,\rvX$ we have $\rvz-\rvmu\not\in\Span\set{\rvX-\rvmu,\rvy-\rvmu}$ with probability $1$ over the randomness of $\rvz$. Thus, $\Pr[T_{\rvX_{-\rvi},\rvmu}(\rvz)=1]=0$.

\paragraph{Bounding the accuracy.} We use the definition of $(\eps,\delta)$-DP to bound the accuracy of $\cM$. 
 \begin{align*}
       \frac{\Pr[G]}{n}
        &\leq \Pr\left[T_{\rvX_{-\rvi},\rvmu}(\rvx_{\rvi})=1\right]\\
        &\leq e^{\eps}\Pr[T_{\rvX_{-\rvi},\rvmu}(\rvx_{\rvi})=1] + \delta && \text{using } T(\rvz)\approx_{(\eps,\delta)}T(\rvx_{\rvi})\\
        & = \delta.
\end{align*}
Rearranging terms, we see that $\Pr[d_{\rvSigma}(\rvy-\rvmu)<\infty]\leq n\delta$.
\end{proof}

\appendix

\section{Concentration Results}
\label{app:sec:concentration}

\begin{lemma}[Concentration of the empirical covariance]
\label{lem:random-matrix-concentration}
    Let $\Sigma$ be a rank-$k$ orthogonal projection in $\R^d$, and $\rvX \sim \cN(0,\Sigma)^{\otimes n}$. There is a numerical constant $c>0$, such that, if $n \le c k$, then 
    \[
    \frac{k}{2n}P_{\rvX} \preceq \hat{\Sigma}_{\rvX} \preceq \frac{2k}{n} P_{\rvX}
    \]
    with probability at least $1-2\exp(-ck)$.
\end{lemma}

\begin{proof}[Proof of \Cref{lem:random-matrix-concentration}]
The result is a direct consequence of standard results on the concentration of eigenvalues of matrices with i.i.d. standard gaussian entries. Indeed, the empirical covariance matrix and the projector on $\Span(\rvX)$ write (for $n \le k \le d$)
\[
    \hat{\Sigma}_{\rvX} = \frac{1}{n}\rvX\rvX^\top = \sum_{i\in[n]}\frac{\sigma_i^2(\rvX)}{n} d_i(\rvX)d_i(\rvX)^\top ~~\text{and}~~
    P_{\rvX} = \sum_{i\in[n]}d_i(\rvX)d_i(\rvX)^\top,
\]
with $(d_i(\rvX))_{i\in[n]}\in\R^d$ an orthonormal family, and $(\sigma_i^{2}(\rvX))_{i\in[n]}$ the singular values of $\rvX\rvX^\top$ (and also of $\rvX^\top\rvX$). Since, in the span of $\Sigma$, $\rvX$ is a $k\times n$ matrix with i.i.d. standard gaussian entries, Theorem 4.6.1 in \cite{Vershynin26} states that, with probability at least $1-\beta$, 
\[
    \norm{\frac{1}{k} \rvX^\top \rvX - \id_n}{2} \le C \max\paren{a, a^2} \quad \text{with}\quad a = \sqrt{\frac{n}{k}} + \sqrt{\frac{\log(2/\beta)}{k}},
    \]
    with $C>0$ a universal constant. It follows that 
    \[
    \max(0, 1-C\max(a,a^2))P_{\rvX} \preceq \frac{n}{k}\hat{\Sigma}_{\rvX} \preceq (1+C\max(a,a^2)) P_{\rvX}.
    \]
    A few algebraic manipulations show that for $\beta = 2\exp(-k/(4C)^2)$ and $k > (4C)^2n$ it holds that $C\max(a,a^2)\le 1/2$, which concludes the proof.
\end{proof}

\begin{lemma}[Chernoff for bounded random variables]
    \label{lem:bounded-rv-chernoff}
    Let $\rvr_1,\dots,\rvr_m$ be independent random variables taking values in $[0,R]$. Then for all $\beta>0$
    $$
    \Pr\brackets{\sum \rvr_j\geq e\sum \Ex[\rvr_j] + R\ln(1/\beta)}\leq \beta.
    $$
\end{lemma}
\begin{proof}
    Observe that for any $m$ independent random variables $\rvr_1,\dots,\rvr_m$ taking values in $[0,R]$ we have 
    \begin{align*}
        \Ex\brackets{e^{\sum \rvr_j/R}}
        & =\prod \Ex\brackets{e^{\rvr_j/R}}\leq\prod \Ex\brackets{1+e\rvr_j/R} \leq e^{(e/R)\sum \Ex[\rvr_j]}, 
    \end{align*}
    where the first inequality follows from the fact that for all $t\in[0,1]$ we have $e^t = 1+t\sum_{k\geq 1} t^{k-1}/k!\leq 1+et$. By a Chernoff bound, for all $t>0$ 
    $$
    \Pr\brackets{\sum \rvr_j/R\geq t}\leq e^{(e/R)\sum \Ex[\rvr_j] - t}.
    $$
    Setting $t=(e/R)\sum \Ex[\rvr_j] + \ln(1/\beta)$ and multiplying through by $R$ completes the proof.
\end{proof}
\begin{lemma}[Sum of heavy tailed random variables]
    \label{lem:heavy-tailed-sum}
    Let $\alpha\in(0,1), t_0>1/2$. Let $\rvy$ be a nonnegative random variable such that $\Pr[\rvy\geq t]\leq C t^{-\alpha}$ for all $t\geq t_0$. Let $\rvy_1,\dots,\rvy_d$ be i.i.d.\ copies of $\rvy$. Then there exists $K=K(C,\alpha,t_0)$, such that for all $m\geq 1$,
\[
    \Pr\brackets{\sum_{i=1}^d \rvy_i \geq m d^{1/\alpha}}
    \leq K m^{-\alpha}.
\]

Additionally for $\alpha = 1$, it holds that 
\[
    \Pr\brackets{\sum_{i=1}^d \rvy_i \geq m d}
    \leq \frac{K_1 \log(md)}{m}, 
\]
where $K_1 = (2C + t_0)$.
\end{lemma}

\begin{proof}
Let $A:=\max\{C,t_0^\alpha\}$. Then for all $t>0$, $\Pr[\rvy\geq t]\leq \min(1,At^{-\alpha})$.
Let
\[
    \rvS=\sum_{i=1}^d \rvy_i,
    \qquad
    \rvM=\max_{1\leq i\leq d}\rvy_i,
    \qquad
    x=md^{1/\alpha}.
\]
By the union bound,
\[
    \Pr\brackets{\rvM\geq \frac{x}{2}}
    \leq d\Pr\brackets{\rvy\geq \frac{x}{2}}
    \leq d\min\paren{1, C\paren{\frac{x}{2}}^{-\alpha}}
    =
    \min(d, A 2^\alpha m^{-\alpha})\le A2^\alpha m^{-\alpha}.
\]

Now let $\rvT=\sum_{i=1}^d \rvy_i\Ind\brackets{\rvy_i\leq x/2}
$. If $\rvS\geq x$ and $\rvM<x/2$, then $\rvT=\rvS\geq x$, so
\[
    \Pr[\rvS\geq x] \le \Pr\paren{\rvS\geq x, ~ \rvM < \frac{x}{2}} + \Pr\paren{\rvS\geq x, ~\rvM \ge \frac{x}{2}} \le \Pr[\rvT \ge x] + \Pr[\rvM \ge \frac{x}{2}].
\]
Moreover, for $\alpha < 1$,
\begin{align*}
    \Ex [\rvT]
    & = d\Ex\brackets{\rvy\Ind[\rvy\leq x/2]} 
    \leq d\int_0^{x/2}\Pr[\rvy\geq t]\,dt \\
    & \leq dt_0 + dA\int_{t_0}^{x/2}t^{-\alpha}\,dt 
    = dt_0 + \frac{dA}{1-\alpha}\paren{\paren{\frac{x}{2}}^{1-\alpha} - t_0^{1-\alpha}},
\end{align*}
thus $\Ex[\rvT] \le \frac{dA}{1-\alpha}2^{\alpha-1}x^{1-\alpha}$.
By Markov's inequality,
\[
    \Pr[\rvT\geq x]
    \leq \frac{\Ex [\rvT]}{x}
    \leq \frac{A2^{\alpha-1}}{1-\alpha}dx^{-\alpha}
    \leq
     \frac{A2^{\alpha-1}}{1-\alpha}m^{-\alpha}.
\]
where the last inequality holds by the fact that $x=md^{1/\alpha}$, so $dx^{-\alpha}=m^{-\alpha}$. 
Combining the two:
\[
    \Pr[\rvS\geq m d^{1/\alpha}]
    \leq
    A2^\alpha m^{-\alpha} +  \frac{A 2^{\alpha-1}}{1-\alpha}m^{-\alpha} \le K_{\alpha,t_0,C}m^{-\alpha}
\]
where $K:= \max\{C,t_0^\alpha\}\left(2^\alpha+\frac{2^{\alpha-1}}{1-\alpha}\right).$

For $\alpha = 1$, 
\[
\Ex [\rvT]
    \leq dt_0 + dC\int_{t_0}^{x/2}t^{-\alpha}\,dt 
    = dt_0 + dC\log\paren{\frac{x}{2t_0}}.
\]
Thus \(\Pr[\rvT\ge x] \le \frac{t_0}{m} + \frac{C\log(md/2t_0)}{m}\)
And 
\[
\Pr[\rvS \ge md^{1/\alpha}] \le \frac{2C}{m}+\frac{t_0}{m}+\frac{C\log(md/2t_0)}{m} \le K_1 \frac{\log(md)}{m}
\]
where $K_1 = (3C + t_0)$.
\end{proof}

\begin{lemma}[Top-$k$ samples from $\invchi(1,1)$]
    \label{lem:top-k}
    Let $\rvsigma^2_1,\dots,\rvsigma^2_d \sim\invchi(1,1)^{\otimes d}$, then there exists an absolute constant $c>0$ such that or all $1\le k \le cd$, 
    \[
    \Pr\brackets{\rvsigma^2_{(k)}\ge \frac{d^2}{4k^2}} \ge 1 - \frac{2}{k}.
    \]
\end{lemma}

\begin{proof}
Let $C'>0$ be a sufficiently large absolute constant. For all $i\in[d]$, since $\rvsigma^2_i \sim \invchi(1,1)$, we have for $t\ge C'$ that $\Pr[\rvsigma^2_i \ge t] = C(t)t^{-1/2}$, with $C(t) \in [1,\sqrt{e}]$. Consider the events $H_i := \set{\rvsigma^2_i\geq d^2/(2k)^2}$ and let $\rvS=\sum_{i \in [d]} \Ind_{H_i}$. It holds that $\Pr[H_i] =  2Ck/d$, and thus $\Ex[\rvS] = 2Ck$. Since $\rvS$ is the sum of independent Bernoulli random variables, Chebyshev's inequality yields 
\begin{align*}
\Pr\brackets{\rvS \le \frac{1}{2} \Ex[\rvS]} 
& \leq \Pr\brackets{\abs{\rvS - \Ex[\rvS]} \ge \frac{1}{2} \Ex[\rvS]} 
\leq \frac{4\Var[\rvS]}{\Ex[\rvS]^2}\\
& \le \frac{4d\Pr[H_1]}{\Ex[\rvS]^2}
= \frac{4}{\Ex[\rvS]}
\leq \frac{2}{k}.
\end{align*}
Thus, at least $k$ of the $\rvsigma^2_{j}$'s are at least $\frac{d^2}{4k^2}$ with probability at least $1-\frac{2}{k}$.
\end{proof}

\begin{lemma}[Concentration of the $\chi^2_{k}$ distribution \citep{LaurentM00}]
\label{lem:laurent-massart}
    Let $\rvx \sim \chi^2_{k}$, then for any $t\ge 0$
    \[
    \Pr[\rvx \ge k + 2\sqrt{kt} + 2t] \le \exp(-t) \quad\text{and}\quad \Pr[\rvx \le k - 2\sqrt{kt}] \le \exp(-t).
    \]
\end{lemma}

\section{Folklore Algorithm}
\label{sec:folklore-estimator}

In this section we explain the folklore algorithm for Gaussian mean estimation, which demonstrates the tightness of \Cref{thm:nd:estimation-lower-bound}. The algorithm proceeds in two phases: first, it estimates $\tr(\Sigma)$ using the average norm of half of the samples, and second, it runs friendly-core with clipping radius $\tr(\Sigma)$. By the guarantees of Friendly-core, the resulting output is the empirical mean on half of the data plus Gaussian noise with variance approximately $\tr(\Sigma)/n^2$. 

\begin{theorem}[Folklore algorithm]
    \label{thm:folklore-estimator}
    For all $\eps,\delta,\beta>0$ there exists an $(\eps,\delta)$-DP mechanism $\cM$ such that for all $\mu\in\R^d$ and positive semi-definite $\Sigma\in\R^{d\times d}$ if 
    \[
    n=\Omega\paren{\frac{\log(1/\beta\delta)}{\eps}}
    \]
    and $\rvX=\rvx_1,\dots,\rvx_n\sim\cN(\mu, \Sigma)$ then
    \[
    \norm{\cM(\rvX)  -\mu}{2}\leq O\paren{\sqrt{\frac{\tr(\Sigma) + \|\Sigma\|_{\text{op}}\log(1/\beta)}{n}} + \frac{\sqrt{d\cdot \tr(\Sigma)\log(1/\delta)}\log(n/\beta)}{\eps n}}
    \]
    with probability at least $1-\beta$. 
\end{theorem}

\begin{proof} In order to define the mechanism $\cM$ we will require the following two results. 

We use a slightly modified version of stable-histogram from \cite{KarwaV18}. The original version of the statement in \Cref{lem:stable-histogram} is from \cite{BrownGSUZ21}, we instantiate it here with bins $\set{[2^{b-1},2^b]}_{b\in\mathbb{Z}}$ and output $\tilde Y=2^{\tilde b}\in\set{2^{b-1},2^b,2^{b+1}}$ instead of $\tilde{b}\in\set{b-1,b,b+1}$.

\begin{lemma}[Stable-histogram \cite{BrownGSUZ21}]
    \label{lem:stable-histogram}
    There exists an algorithm $\mathrm{StableHistogram}_{\eps,\delta}$ that is $(\eps,\delta)$-differentially private and satisfies the following guarantee:  
    Let $\rvz_1,\ldots,\rvz_n$ be drawn i.i.d.\ from distribution $P$. 
    Suppose that there exists $b\in\mathbb{Z}$ and a constant $\beta'<\frac{1}{4}$, such that $\Pr[\rvz_i\notin [2^{b-1},2^{b+1}]]\leq \beta'$ for any fixed $i\in[n]$. 
    There exists a constant $C>0$ such that, for all $0<\eps,\beta,\delta<1$, if \[n\geq\frac{C}{\eps}\log\frac{1}{\beta\delta},\] then $\mathrm{StableHistogram}_{\eps,\delta}(\rvz_1,\ldots,\rvz_n)\in [2^{b-1},2^{b+1}]$ with probability at least $1-\beta$.
\end{lemma}

We will also use the friendly-core aggregation algorithm discovered by \cite{TsfadiaCKMS22} to estimate the mean. The exact statement of \Cref{lem:friendly-core} does not appear explicitly in \cite{TsfadiaCKMS22}, but it follows immediately from their results.

\begin{lemma}[Friendly-core averaging]
    \label{lem:friendly-core}
    Fix $\eps,\delta,R>0$. There exists an $(\eps,\delta)$-DP algorithm $\cA$ that gets as input $X=x_1,\dots,x_n$ and satisfies the following:
    If $n\geq \tilde\Omega\paren{\sqrt{\log(1/\beta\delta)}/\eps}$ and $\norm{x_i-x_j}{2}\leq R$ for all $i,j\in[n]$ then 
    \[
    \norm{\cA(X) - \frac{1}{n}\sum x_i}{2}\leq O\paren{\frac{R\sqrt{d\log(1/\delta)\log(1/\beta)}}{\eps n}}
    \]
    with probability at least $1-\beta$.
\end{lemma}

\begin{proof}[Proof of \Cref{lem:friendly-core}]
    Let $\cA$ denote the algorithm \textsf{FC\textunderscore Avg} from \cite{TsfadiaCKMS22} run with zCDP parameters $\rho= \eps^2/10\log(1/\delta)$ and $\delta_0=\delta/2$ and diameter $R$. 
    
    First we argue that $\cA$ is private. By Theorem 5.2 of \cite{TsfadiaCKMS22}, algorithm $\cA$ is $(\rho,\delta)$-zCDP, which, by Fact 2.7 of \cite{TsfadiaCKMS22} with $\delta'\gets\delta_0$, implies that $\cA$ is $(\rho + 2\sqrt{\rho\ln(1/\delta_0)}, 2\delta_0)$-DP. By our setting of $\rho$ and $\delta_0$ algorithm $\cA$ is $(\eps,\delta)$-DP.  

    To see that $\cA$ is accurate, we can apply Claim C.4 of \cite{TsfadiaCKMS22}, which states that if $n=\Omega(\sqrt{\ln(1/\rho\beta\delta_0)/\rho})$ and $\norm{x_i-x_j}{2}\leq R$ for all $i,j\in[n]$ then
    \[
    \norm{\cA(X) - \frac{1}{n}\sum x_i}{2}\leq O\paren{\frac{R}{n}\sqrt{\frac{d\log(1/\beta)}{\rho}}}
    \]
    with probability at least $1-\beta$. Using our setting of $\rho$ and $\delta_0$ concludes the proof.\\
    
    \par\noindent{\footnotesize \textit{Note: The requirement on $n$ in Claim C.4 of \cite{TsfadiaCKMS22} is missing a square root---it reads $n\geq \log(1/\beta\delta)/\rho$; however, the proof of Claim C.4 only requires $n\geq \sqrt{\log(1/\beta\delta)/\rho}$ in order to invoke the relevant claim (Claim C.3), and thus the requirement on $n$ in our proof is sufficient.}}
\end{proof}

Next, we define the mechanism $\cM$. Let $C>0$ be a sufficiently large absolute constant. 

\begin{algorithm}[H]
    \begin{algorithmic}
        \caption{\label{alg:folklore-estimator} Folklore estimator $\cM$}
        \STATE \textbf{Input:} Privacy parameter $\eps,\delta\in(0,1)$, dataset $X=x_1,\dots,x_n$, failure probability $\beta>0$. 
        \STATE \textbf{Output:} Estimate $\tilde{\mu}\in\R^d$
        \STATE Set $\ell\gets \frac{C}{\eps}\log(1/\beta\delta)$
        \STATE Partition $X$ into $X_0$ and $X_1$, and $X_0$ into $B_1,\dots, B_\ell$
        \STATE For each $i\in [\ell]$ and $x_1,\dots,x_{n/2\ell}\in B_i$ let $y_j=\frac{1}{\sqrt{2}}\paren{x_{2j-1}-x_{2j}}$ for each $j\in[n/4\ell]$.
        \STATE For each $i\in[\ell]$ let 
        
        $$\bar y_i\gets \frac{4\ell}{n}\sum_{y_j\in B_i} \norm{y_j}{2}^2$$
        
        \STATE Let $\tilde \rvy$ be the output of $\mathrm{StableHistogram}_{\eps/2,\delta/2}$ on input $(\bar y_i)_{i\in [\ell]}$
        \STATE Let $R\gets C\sqrt{\tilde \rvy\log(n/\beta)}$
        \STATE Output $\tilde{\rvmu}\gets \cA_{\eps/2,\delta/2}(X_1,R)$ where $\cA$ is the algorithm from \Cref{lem:friendly-core} \end{algorithmic}
\end{algorithm}

First, we argue that $\cM$ is private. Since stable-histogram and $\cA$ are $(\eps/2,\delta/2)$-DP, basic composition and postprocessing yields that $\cM$ is $(\eps,\delta)$-DP. 

Next, we show that the accuracy guarantee holds. The basic idea is to show that each $\bar \rvy_i$ gives a multiplicative approximation to $\tr(\Sigma)$ with high probability, and thus, stable-histogram returns an estimate $\tilde{\rvy}$ that is a multiplicative approximation to $\tr(\Sigma)$. We then argue that with high probability $\norm{\rvx_i-\rvx_j}{2}\leq R$ for all $\rvx_i$ and $\rvx_j$ in $\rvX_1$. Applying the guarantee of \Cref{lem:friendly-core} and a union bound suffices to complete the argument. The remainder of the proof formalizes this intuitive argument.

\begin{lemma}[Concentration of the norm]
    \label{lem:norm-concentration}
    Let $\rvY=\rvy_1,\dots,\rvy_m\sim\cN(0,\Sigma)$ then for all $\beta>0$
    \[
    \abs{\frac{1}{m}\sum \norm{\rvy_i}{2}^2 - \tr(\Sigma)} \leq \tr(\Sigma)\sqrt{\frac{\log(2/\beta)}{m}} + \|\Sigma\|_{\text{op}}\frac{\log(2/\beta)}{m}\]
    with probability at least $1-\beta$.
\end{lemma}
\begin{proof}
    This follows from rotation invariance of $\cN(0,\Sigma)$ and Chebyshev's inequality. Observe that if $\rvy\sim\cN(0,\Sigma)$, then we can equivalently write $\rvy$ as $\Sigma^{1/2}\rvz$ where $\rvz\sim\cN(0,\id)$. Now, if $\Sigma=V D V^\top$ then $\norm{\rvy}{2}^2 = \rvz^\top V D V^\top \rvz$. By rotation invariance of the Gaussian distribution, the random variable $\tilde{\rvz}=V^\top \rvz$ is still distributed according to $\cN(0,\id)$. Thus, $\norm{\rvy}{2}^2=\tilde{\rvz}^\top D\tilde{\rvz}=\sum_{j\in[d]} \lambda_{j}\rvz[j]^2$, where $\lambda_j$ is the $\ord{j}$ eigenvalue of $\Sigma$ and $\rvz[j]\sim\chi^2_1$. 

    Applying the above argument to each $\rvy_i$, we have that 
    $$
    \frac{1}{m}\sum_{i\in[m]} \|\rvy_i\|^2_2 = \frac{1}{m}\sum_{\substack{i\in[m] \\ j\in[d]}} \lambda_j\rvz_i[j]^2
    $$
    
    Since $\Ex[\|\rvy_i\|^2_2]=\tr(\Sigma)$, and since the $\rvz_i[j]^2$'s are independent for each $i$ and $j$, we can apply Scalar Bernstein (e.g., Corollary 2.9.2 \cite{Vershynin26}) to see that 
    $$
    \Pr\left[\abs{\frac{1}{m}\sum_{\substack{i\in[m] \\ j\in[d]}}\lambda_j\rvz_i[j]^2-\tr(\Sigma)}\geq \tr(\Sigma)\sqrt{\frac{\log(2/\beta)}{m}} + \|\Sigma\|_{\text{op}}\frac{\log(2/\beta)}{m} \right]\leq \beta.\qedhere
    $$    
\end{proof}

By our setting of $n$ and $\ell$, and by \Cref{lem:norm-concentration} with $m=n/4\ell$ and $\beta\gets 1/5$, we have that $\Pr[\bar \rvy_i\not\in \tr(\Sigma)\cdot[1/2,2]]\leq 1/5$. Thus, by \Cref{lem:stable-histogram}, and our setting of $n$, we have that $\tilde{\rvy}\in \tr(\Sigma)\cdot[1/2,2]$ with probability at least $1-\beta$. 

Let $B_0$ be the event that $\tilde{\rvy}$ is not in this interval, and let $B_1$ be the event that some $\rvx_i-\mu$ has $\norm{\rvx_i-\mu}{2}\geq C\sqrt{\tr(\Sigma)\log(n/\beta)}$ for some large constant $C>0$. By \Cref{lem:norm-concentration} with $m=1$ and the union bound over the $n$ datapoints, we have that $\Pr[B_1]\leq \beta$. Now, conditioned on $\overline{B_0\cup B_1}$, we have that $\norm{\rvx_i-\rvx_j}{2}\leq R$ for all $i$ and $j$, and thus, by \Cref{lem:friendly-core}, 
\[
\norm{\cA(\rvX) - \frac{2}{n}\sum_{\rvx_i\in X_1} \rvx_i}{2}\leq O\paren{\frac{R\sqrt{d\log(1/\delta)\log(1/\beta)}}{\eps n}}
\]
with probability at least $1-\beta$. Let $B_2$ be the event that $\tilde\mu$ fails to satisfy the above inequality. Then 
\[
\Pr[B_2]=\Pr[B_2\mid B_1\cup B_0]\Pr[B_1\cup B_0] + \Pr[B_2\mid \overline{B_1\cup B_0}]\Pr[\overline{B_1\cup B_0}]\leq 3\beta.
\]
To complete the argument, we need to show that the empirical mean on $X_1$ is close to $\mu$ with high probability. Let $B_3$ be the event that 
\[
\norm{\frac{2}{n}\sum_{\rvx_i\in X_1} \rvx_i - \mu}{2}\geq C\sqrt{\frac{\tr(\Sigma)+\|\Sigma\|_{\text{op}}\log(1/\beta)}{n}}.
\]
Standard arguments for the min-max rate of Gaussian mean estimation imply that $\Pr[B_3]\leq \beta$. Since $\tilde\rvmu$ satisfies the accuracy guarantee of \Cref{thm:folklore-estimator} whenever neither of $B_2$ or $B_3$ occur, the union bound over $B_2$ and $B_3$ implies that the guarantee holds with probability at least $1-\tilde\beta$ with $\tilde \beta = 4\beta$.    
\end{proof}


\begin{thebibliography}{BGSUZ21}
\bibitem[AAK21]{AdenAliAK21}
Ishaq Aden-Ali, Hassan Ashtiani, and Gautam Kamath. “On the Sample
Complexity of Privately Learning Unbounded High-Dimensional {G}aussians.”
\emph{International Conference on Algorithmic Learning Theory}. ALT 2021.
\bibitem[ALNP24]{AumullerLNP23}
Martin Aumüller, Christian Janos Lebeda, Boel Nelson, and Rasmus Pagh.
“{PLAN}: Variance-Aware Private Mean Estimation.” \emph{Privacy Enhancing
Technologies Symposium}. PETS 2024. \href
{https://arxiv.org/abs/2306.08745} {\texttt{(Link)}}.
\bibitem[BBFST21]{BrownBFST21}
Gavin Brown, Mark Bun, Vitaly Feldman, Adam Smith, and Kunal Talwar.
“When is Memorization of Irrelevant Training Data Necessary for
High-Accuracy Learning?” \emph{arXiv preprint arXiv:2012.06421} (2021).
\href {https://arxiv.org/abs/2012.06421} {\texttt{(Link)}}.
\bibitem[BGSUZ21]{BrownGSUZ21}
Gavin Brown, Marco Gaboardi, Adam Smith, Jonathan Ullman, and
Lydia Zakynthinou. “Covariance-Aware Private Mean Estimation Without
Private Covariance Estimation.” \emph{Advances in Neural Information
Processing Systems}. NeurIPS 2021. \href
{https://arxiv.org/abs/2106.13329} {\texttt{(Link)}}.
\bibitem[BHS23]{BrownHS23}
Gavin Brown, Samuel B. Hopkins, and Adam Smith. “Fast, Sample-Efficient,
Affine-Invariant Private Mean and Covariance Estimation for Subgaussian
Distributions.” \emph{Conference on Learning Theory}. COLT 2023.
\bibitem[BKSW19]{BunKSW19}
Mark Bun, Gautam Kamath, Thomas Steinke, and Zhiwei Steven Wu. “Private
Hypothesis Selection.” \emph{Advances in Neural Information Processing
Systems}. NeurIPS 2019.
\bibitem[BUV14]{BunUV14}
Mark Bun, Jonathan Ullman, and Salil Vadhan. “Fingerprinting Codes and
the Price of Approximate Differential Privacy.” \emph{ACM Symposium on
Theory of Computing}. STOC 2014. \href {https://arxiv.org/abs/1311.3158}
{\texttt{(Link)}}.
\bibitem[Car+21]{Carlini+21}
Nicholas Carlini, Florian Tramer, Eric Wallace, Matthew Jagielski,
Ariel Herbert-Voss, Katherine Lee, Adam Roberts, Tom Brown, et al.
“Extracting training data from large language models.” \emph{USENIX
Security Symposium}. USENIX Security 2021. \href
{https://arxiv.org/abs/2012.07805} {\texttt{(Link)}}.
\bibitem[CDMM26]{ChenDMM26}
Sitan Chen, Jingqiu Ding, Mahbod Majid, and Walter McKelvie.
“Computation-Utility-Privacy Tradeoffs in Bayesian Estimation.” \emph{ACM
Symposium on Theory of Computing}. STOC 2026. \href
{https://doi.org/10.1145/3798129.3800869} {\texttt{(Link)}}.
\bibitem[CLEKS19]{CarliniLEKS19}
Nicholas Carlini, Chang Liu, Úlfar Erlingsson, Jernej Kos, and Dawn Song.
“The secret sharer: Evaluating and testing unintended memorization in
neural networks.” \emph{USENIX Security Symposium}. USENIX Security 2019.
\href {https://arxiv.org/abs/1802.08232} {\texttt{(Link)}}.
\bibitem[CWZ19]{CaiWZ19}
T. Tony Cai, Yichen Wang, and Linjun Zhang. “The Cost of Privacy: Optimal
Rates of Convergence for Parameter Estimation with Differential Privacy.”
\emph{arXiv preprint arXiv:1902.04495} (2019). \href
{https://arxiv.org/abs/1902.04495} {\texttt{(Link)}}.
\bibitem[CWZ23]{CaiWZ23}
T. Tony Cai, Yichen Wang, and Linjun Zhang. “Score attack: A lower bound
technique for optimal differentially private learning.” \emph{arXiv
preprint arXiv:2303.07152} (2023). \href
{https://arxiv.org/abs/2303.07152} {\texttt{(Link)}}.
\bibitem[DJYZZ24]{DaganJYZZ24}
Yuval Dagan, Michael Jordan, Xuelin Yang, Lydia Zakynthinou, and
Nikita Zhivotovskiy. “Dimension-free private mean estimation for
anisotropic distributions.” \emph{Advances in Neural Information
Processing Systems}. NeurIPS 2024. \href
{https://arxiv.org/abs/2411.00775} {\texttt{(Link)}}.
\bibitem[DL22]{depersin2022optimal}
Jules Depersin and Guillaume Lecué. “Optimal robust mean and location
estimation via convex programs with respect to any pseudo-norms.”
\emph{Probability Theory and Related Fields} 183.3 (2022).
\bibitem[DMNS06]{DworkMNS06}
Cynthia Dwork, Frank McSherry, Kobbi Nissim, and Adam Smith. “Calibrating
Noise to Sensitivity in Private Data Analysis.” \emph{Conference on
Theory of Cryptography}. TCC 2006. \href
{https://www.iacr.org/archive/tcc2006/38760266/38760266.pdf}
{\texttt{(Link)}}.
\bibitem[DMT07]{DworkMT07}
Cynthia Dwork, Frank McSherry, and Kunal Talwar. “The price of privacy
and the limits of {LP} decoding.” \emph{ACM Symposium on Theory of
Computing}. STOC 2007. \href {https://doi.org/10.1145/1250790.1250804}
{\texttt{(Link)}}.
\bibitem[DN03]{DinurN03}
Irit Dinur and Kobbi Nissim. “Revealing information while preserving
privacy.” \emph{ACM Symposium on Principles of Database Systems}. PODS
2003. \href {https://doi.org/10.1145/773153.773173} {\texttt{(Link)}}.
\bibitem[DSSUV15]{DworkSSUV15}
Cynthia Dwork, Adam Smith, Thomas Steinke, Jonathan Ullman, and
Salil Vadhan. “Robust Traceability from Trace Amounts.” \emph{IEEE
Symposium on Foundations of Computer Science}. FOCS 2015. \href
{https://doi.org/10.1109/FOCS.2015.46} {\texttt{(Link)}}.
\bibitem[DY08]{DworkY08}
Cynthia Dwork and Sergey Yekhanin. “New efficient attacks on statistical
disclosure control mechanisms.” \emph{International Cryptology
Conference}. CRYPTO 2008. \href
{https://iacr.org/archive/crypto2008/51570469/51570469.pdf}
{\texttt{(Link)}}.
\bibitem[Fel20]{Feldman20}
Vitaly Feldman. “Does learning require memorization? {A} short tale about
a long tail.” \emph{ACM Symposium on Theory of Computing}. STOC 2020.
\href {https://arxiv.org/abs/1906.05271} {\texttt{(Link)}}.
\bibitem[GLSUZ26]{GaucherLSUZ26}
Renaud Gaucher, Ephraim Linder, Adam Smith, Jonathan Ullman, and
Lydia Zakynthinou. \emph{A Simple Presentation of Robust Membership
Inference Attacks}. 2026. arXiv: \href {https://arxiv.org/abs/2610.36458}
{\nolinkurl {2610.36458} \texttt{[cs.CR]}}. \href
{https://arxiv.org/abs/2610.36458} {\texttt{(Link)}}.
\bibitem[Hom+08]{Homer+08}
Nils Homer, Szabolcs Szelinger, Margot Redman, David Duggan,
Waibhav Tembe, Jill Muehling, John V. Pearson, Dietrich A. Stephan,
et al. “Resolving individuals contributing trace amounts of {DNA} to
highly complex mixtures using high-density {SNP} genotyping microarrays.”
\emph{PLoS Genetics} 4.8 (2008). \href
{https://journals.plos.org/plosgenetics/article?id=10.1371/journal.pgen.1000167}
{\texttt{(Link)}}.
\bibitem[KDH23]{KuditipudiDH23}
Rohith Kuditipudi, John C. Duchi, and Saminul Haque. “A Pretty Fast
Algorithm for Adaptive Private Mean Estimation.” \emph{Conference on
Learning Theory}. COLT 2023.
\bibitem[KLSU19]{KamathLSU19}
Gautam Kamath, Jerry Li, Vikrant Singhal, and Jonathan Ullman. “Privately
Learning High Dimensional Distributions.” \emph{Conference on Learning
Theory}. COLT 2019. \href {https://arxiv.org/abs/1805.00216}
{\texttt{(Link)}}.
\bibitem[KMS22]{KamathMS22}
Gautam Kamath, Argyris Mouzakis, and Vikrant Singhal. “New Lower Bounds
for Private Estimation and a Generalized Fingerprinting Lemma.”
\emph{Advances in Neural Information Processing Systems}. NeurIPS 2022.
\bibitem[KPST26]{KumarPST26}
Syamantak Kumar, Shourya Pandey, Purnamrita Sarkar, and Kevin Tian. “On
the Curse of Dimensionality in Private Sparse Covariance Estimation and
{PCA}.” \emph{Conference on Learning Theory}. COLT 2026. \href
{https://proceedings.mlr.press/v336/kumar26b.html} {\texttt{(Link)}}.
\bibitem[KRSU10]{KasiviswanathanRSU10}
Shiva Prasad Kasiviswanathan, Mark Rudelson, Adam Smith, and
Jonathan Ullman. “The price of privately releasing contingency tables and
the spectra of random matrices with correlated rows.” \emph{ACM Symposium
on Theory of Computing}. STOC 2010. \href
{https://doi.org/10.1145/1806689.1806795} {\texttt{(Link)}}.
\bibitem[KV18]{KarwaV18}
Vishesh Karwa and Salil Vadhan. “Finite Sample Differentially Private
Confidence Intervals.” \emph{Conference on Innovations in Theoretical
Computer Science}. ITCS 2018. \href {https://arxiv.org/abs/1711.03908}
{\texttt{(Link)}}.
\bibitem[LKKO21]{LiuKKO21}
Xiyang Liu, Weihao Kong, Sham Kakade, and Sewoong Oh. “Robust and
differentially private mean estimation.” \emph{Advances in Neural
Information Processing Systems}. NeurIPS 2021.
\bibitem[LM00]{LaurentM00}
Béatrice Laurent and Pascal Massart. “Adaptive estimation of a quadratic
functional by model selection.” \emph{The Annals of Statistics} 28.5
(2000). \href {https://doi.org/10.1214/aos/1015957395} {\texttt{(Link)}}.
\bibitem[MZ25]{MinasyanZ23}
Arshak Minasyan and Nikita Zhivotovskiy. “Statistically Optimal Robust
Mean and Covariance Estimation for Anisotropic {G}aussians.”
\emph{Mathematical Statistics and Learning}. Vol. 8. 2025. \href
{https://arxiv.org/abs/2301.09024} {\texttt{(Link)}}.
\bibitem[Nar25]{narayanan25better}
Shyam Narayanan. “Better and Simpler Lower Bounds for Differentially
Private Statistical Estimation.” \emph{IEEE Trans. Inf. Theor.} 71.2
(Feb. 2025). \textsc{issn}: 0018-9448. \textsc{doi}: \href
{https://doi.org/10.1109/TIT.2024.3511624} {\nolinkurl
{10.1109/TIT.2024.3511624}}. \href
{https://doi.org/10.1109/TIT.2024.3511624} {\texttt{(Link)}}.
\bibitem[PH25]{PortellaH25}
Victor S. Portella and Nicholas J. A. Harvey. “Lower Bounds for Private
Estimation of {Gaussian} Covariance Matrices under All Reasonable
Parameter Regimes.” \emph{Conference on Learning Theory}. COLT 2025.
\href {https://arxiv.org/abs/2404.17714} {\texttt{(Link)}}.
\bibitem[PTU24]{PeterTU24}
Naty Peter, Eliad Tsfadia, and Jonathan Ullman. “Smooth Lower Bounds for
Differentially Private Algorithms via Padding-and-Permuting
Fingerprinting Codes.” \emph{Conference on Learning Theory}. COLT 2024.
\href {https://arxiv.org/abs/2307.07604} {\texttt{(Link)}}.
\bibitem[Smi11]{Smith11}
Adam Smith. “Privacy-Preserving Statistical Estimation with Optimal
Convergence Rates.” \emph{ACM Symposium on Theory of Computing}. STOC
2011.
\bibitem[SOJH09]{SankararamanOJH09}
Sriram Sankararaman, Guillaume Obozinski, Michael I. Jordan, and
Eran Halperin. “Genomic privacy and limits of individual detection in a
pool.” \emph{Nature Genetics} 41.9 (2009). \href
{https://www.nature.com/articles/ng.436} {\texttt{(Link)}}.
\bibitem[SS21]{SinghalS21}
Vikrant Singhal and Thomas Steinke. “Privately Learning Subspaces.”
\emph{Advances in Neural Information Processing Systems}. NeurIPS 2021.
\bibitem[SSSS17]{ShokriSSS17}
Reza Shokri, Marco Stronati, Congzheng Song, and Vitaly Shmatikov.
“Membership Inference Attacks against Machine Learning Models.”
\emph{IEEE Symposium on Security \& Privacy}. IEEE S\&P 2017. \href
{https://arxiv.org/abs/1610.05820} {\texttt{(Link)}}.
\bibitem[SU17]{SteinkeU17}
Thomas Steinke and Jonathan Ullman. “Tight Lower Bounds for
Differentially Private Selection.” \emph{IEEE Symposium on Foundations of
Computer Science}. FOCS 2017.
\bibitem[TCKMS22]{TsfadiaCKMS22}
Eliad Tsfadia, Edith Cohen, Haim Kaplan, Yishay Mansour, and Uri Stemmer.
“{F}riendly{C}ore: Practical Differentially Private Aggregation.”
\emph{International Conference on Machine Learning}. ICML 2022. \href
{https://arxiv.org/abs/2110.10132} {\texttt{(Link)}}.
\bibitem[Tsf24]{Tsfadia24}
Eliad Tsfadia. “On differentially private subspace estimation in a
distribution-free setting.” \emph{Advances in Neural Information
Processing Systems}. NeurIPS 2024.
\bibitem[Ver26]{Vershynin26}
Roman Vershynin. \emph{High-Dimensional Probability: An Introduction with
Applications in Data Science}. 2nd ed. Cambridge Series in Statistical
and Probabilistic Mathematics. Cambridge University Press, 2026.
\bibitem[YGFJ18]{YeomGFJ18}
Samuel Yeom, Irene Giacomelli, Matt Fredrikson, and Somesh Jha. “Privacy
risk in machine learning: Analyzing the connection to overfitting.”
\emph{IEEE Computer Security Foundations Symposium}. CSF 2018. \href
{https://arxiv.org/abs/1709.01604} {\texttt{(Link)}}.
\end{thebibliography}
\end{document}